\pdfoutput=1

\documentclass[11pt]{article}

\usepackage[final]{acl}

\usepackage{times}
\usepackage{latexsym}

\usepackage[T1]{fontenc}

\usepackage[utf8]{inputenc}

\usepackage{microtype}

\usepackage{inconsolata}

\usepackage{graphicx}

\usepackage{booktabs}
\usepackage{multirow}
\usepackage{graphicx}
\usepackage{amsmath}
\usepackage{lipsum}

\usepackage{CJKutf8}
\usepackage{microtype}
\usepackage{hyperref}
\usepackage{url}
\usepackage{booktabs}
\usepackage{amssymb}
\usepackage{float}

\usepackage[most]{tcolorbox}
\usepackage{caption}
\usepackage{fancyvrb}

\usepackage{makecell}

\newcommand{\coloredsquare}[1]{\textcolor{#1}{$\blacksquare$}}

\definecolor{CustomBlue}{RGB}{57,83,191}

\definecolor{cp1_color}{HTML}{4285f4} 
\definecolor{cp2_color}{HTML}{FF5722}

\usepackage{amssymb}
\usepackage{arydshln}
\makeatletter
\def\adl@drawiv#1#2#3{%
        \hskip.5\tabcolsep
        \xleaders#3{#2.5\@tempdimb #1{1}#2.5\@tempdimb}%
                #2\z@ plus1fil minus1fil\relax
        \hskip.5\tabcolsep}
\newcommand{\cdashlinelr}[1]{%
  \noalign{\vskip 1.3pt
           \global\let\@dashdrawstore\adl@draw
           \global\let\adl@draw\adl@drawiv}
  \cdashline{#1}[.4pt/2pt]
  \noalign{\global\let\adl@draw\@dashdrawstore
           \vskip 3pt}}
\makeatother

\definecolor{custom1}{RGB}{15,157,88}

\newcommand{\flores}{\textsc{Flores-200}}
\newcommand{\wmtplusplus}{\textsc{WMT24++}}
\newcommand{\ntrex}{\textsc{NTREX}}
\newcommand{\bleu}{\textsc{BLEU}}
\newcommand{\comet}{\textsc{COMET}}

\newcommand{\towerblocks}{\textsc{TowerBlocks}}
\newcommand{\europarl}{\textsc{EuroParl}}
\newcommand{\newscommentary}{\textsc{NewsCommentary}}

\newcommand{\cometkiwi}{\textsc{COMET-}\textsc{\scriptsize Kiwi}}

\newcommand{\salamandraTA}{\textsc{Salamandra\textbf{TA}}}
\newcommand{\salamandra}{\textsc{Salamandra}}
\newcommand{\salamandraTAversiontwo}{\textsc{Salamandra\textbf{TA}}{\scriptsize \textsc{-v2}}}

\newcommand{\salamandraTAcpone}{\textbf{ \textsc{CPT-v1}}}
\newcommand{\salamandraTAcptwo}{\textbf{ \textsc{CPT-v2}}}

\newcommand{\salamandraTAitone}{\textbf{ \textsc{IT-v1}}}
\newcommand{\salamandraTAittwo}{\textbf{ \textsc{IT-v2}}}

\newcommand{\salamandraTAsevenbase}{\textsc{Salamandra\textbf{TA}{\normalsize 7B-Base}}}
\newcommand{\salamandraTAtwobase}{\textsc{Salamandra\textbf{TA}{\normalsize 2B-Base}}}

\title{From \salamandra\ to \textsc{Salamandra\textcolor{custom1}{ \textbf{TA}}}: \\ {\large BSC Submission for WMT25 General Machine Translation Shared Task} }

\author{
Javier Garcia Gilabert\thanks{Core Contributor.   }$^{1}$ \;\; Xixian Liao\footnotemark[1]$^{1}$ \;\; Severino Da Dalt$^{1}$ \;\; Ella Bohman$^{1}$ \\
\textbf{Audrey Mash$^{1}$ \;\; Francesca De Luca Fornaciari$^{1}$ \;\; Irene Baucells$^{1}$ \;\; Joan Llop$^{1}$} \\
\textbf{Miguel Claramunt Argote$^{1}$ \;\; Carlos Escolano$^{1,2}$ \;\; Maite Melero$^{1}$} \\
$^{1}$Barcelona Supercomputing Center \\
$^{2}$Universitat Politècnica de Catalunya
}

\begin{document}
\maketitle
\begin{abstract}

In this paper, we present the \salamandraTA\ family of models, an improved iteration of \salamandra\ LLMs \cite{gonzalezagirre2025salamandratechnicalreport} specifically trained to achieve strong performance in translation-related tasks for 38 European languages. \salamandraTA\ comes in two scales: 2B and 7B parameters. For both versions, we applied the same training recipe with a first step of continual pre-training on parallel data, and a second step of supervised fine-tuning on high-quality instructions.

The BSC submission to the WMT25 General Machine Translation shared task is based on the 7B variant of \salamandraTA. We first adapted the model vocabulary to support the additional non-European languages included in the task. This was followed by a second phase of continual pre-training and supervised fine-tuning, carefully designed to optimize performance across all translation directions for this year's shared task. For decoding, we employed two quality-aware strategies: Minimum Bayes Risk Decoding and  Tuned Re-ranking using \comet\ and \cometkiwi\ respectively.

We publicly release both the 2B and 7B versions of \salamandraTA, along with the newer \salamandraTAversiontwo\ model, on Hugging Face\footnote{ \href{https://huggingface.co/BSC-LT/salamandraTA-7b-instruct}{ \textsc{Salamandra\textbf{TA}{\normalsize 7B}}{\scriptsize \textsc{-v1}} }, \href{https://huggingface.co/BSC-LT/salamandraTA-2b-instruct}{ \textsc{Salamandra\textbf{TA}{\normalsize 2B}}{\scriptsize \textsc{-v1}} }  and \href{https://huggingface.co/LangTech-MT/salamandraTA-7b-instruct-WMT25}{ \textsc{Salamandra\textbf{TA}{\normalsize 7B}}{\scriptsize \textsc{-v2}} }. }.

\end{abstract}

\begin{figure*}[t]
\centering
  \includegraphics[width=\linewidth]{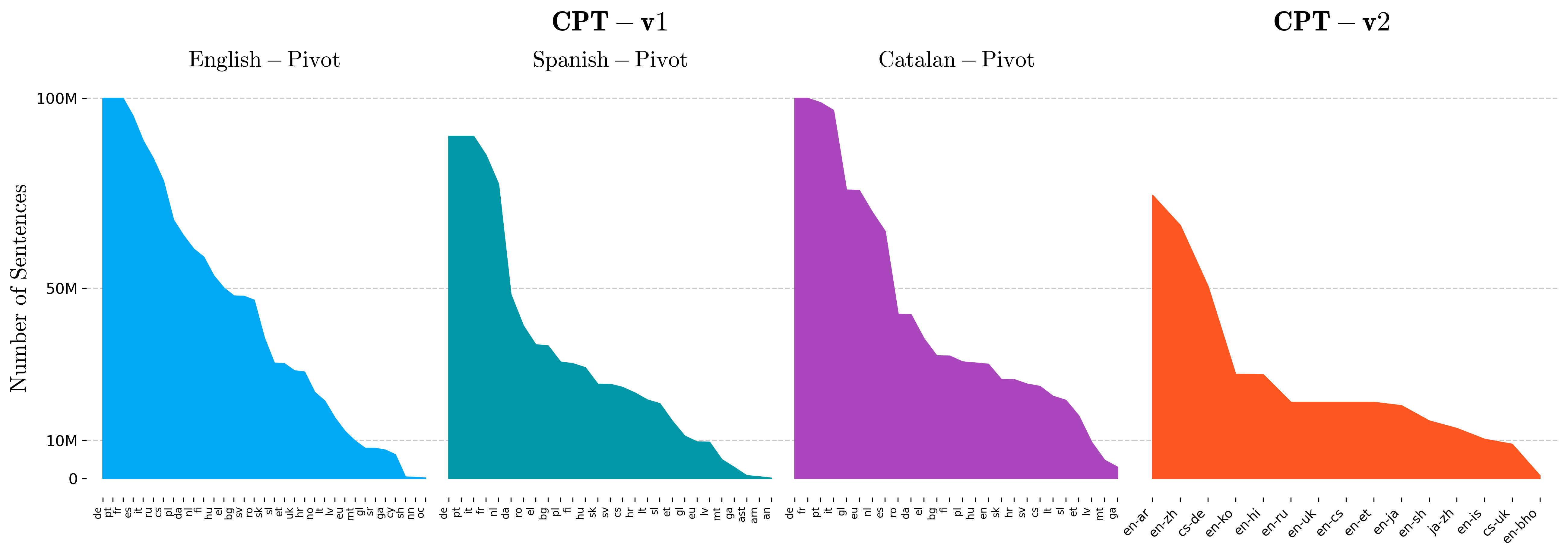}
  \caption {Distribution of sentence pairs for continual pre-training. The first three plots (\coloredsquare{cp1_color}\salamandraTAcpone ) show the number of sentence pairs pivoting in English, Spanish and Catalan, respectively. The fourth plot (\coloredsquare{cp2_color} {\salamandraTAcptwo }) corresponds to the second continual pre-training phase with direct language pairs.}\label{fig:data_distribution}
\end{figure*}

\section{Introduction}

Traditionally, Massively Multilingual Neural Machine Translation (MMNMT) relied on the encoder-decoder architecture to translate across multiple languages \cite{10.5555/3546258.3546365, nllbteam2022language}. More recently, however, Large Language Models (LLMs) have demonstrated strong MMNMT capabilities \cite{zhu-etal-2024-multilingual} and thus some works have proposed several strategies to improve the translation capabilities of a pre-trained LLM model and better align it with human translations \cite{DBLP:conf/icml/0006HB23, alves2024tower, xu2024a}.  

One such approach is continual pre-training using a combination of monolingual and parallel corpora followed by supervised fine-tuning \cite{alves2024tower}. However, most previous approaches have predominantly relied on English-centric parallel corpora. This has been shown to bias the models towards English-centric latent representations \cite{zhang2025exploringtranslationmechanismlarge} which has been attributed to the language distribution used in the training corpora \cite{zhong2024englishcentricllmslanguagemultilingual}. It is well known that training with only a single bridge language can negatively impact translation performance across zero-shot language pairs, due to limited cross-lingual transfer \cite{arivazhagan2019massively}. Unlike previous works, in this paper we rely on parallel corpora only for the continual pre-training stage pivoting on three bridge languages.

When working with pre-trained language models on languages not covered by their original tokenizer, a highly effective solution involves replacing the existing tokenizer with a more comprehensive one that supports such languages. This strategy necessitates randomly initializing the embeddings for the newly introduced tokens. These new embeddings are then rapidly optimized through continual pre-training (CPT). This method has not only proven to be viable but also demonstrably improves the model's overall performance in the target languages, even if the original model was never exposed to data from these languages during its initial training \cite{da-dalt-etal-2024-flor}.

Throughout this paper, we present the \salamandraTA\ family of models, which serve as the backbone models of the BSC team’s submission to the WMT25 General Machine Translation Shared Task. Our participation covers 15 out of the 16 translation directions in the general MT task under the constrained track. Additionally, we took part in the multilingual subtask for 7 out of the 16 directions. Contributions of this work are listed as follows:

\begin{itemize}

    \item While most previous work have relied on English-centric parallel corpora for building translation-focused LLMs, we build \salamandraTA\ pivoting in three languages for continual pre-training; English, Spanish and Catalan across 172 supervised directions.

    \item We show that instruction tuning improves both translation quality and robustness to character-level noise.

    \item We release all model checkpoints to facilitate reproducibility and future research on massively multilingual machine translation.
    
\end{itemize}

\section{Data}

Our base models are \salamandra-2B and \salamandra-7B \cite{gonzalezagirre2025salamandratechnicalreport}, which were trained from scratch on highly multilingual data. However, \salamandra\ models were not exposed to parallel data during pre-training. To address this, and following \citet{alves2024tower}, we improve their multilingual machine translation capabilities by performing continual pre-training on parallel data covering 38 European languages (35 of which were already present in the original pre-training corpus). This step is followed by supervised fine-tuning using high-quality instruction data. In this section, we detail the datasets used for both continual pre-training and supervised fine-tuning.

\begin{table*}[!ht]
\centering
\setlength{\tabcolsep}{4.75pt}
\footnotesize
\renewcommand{\arraystretch}{1.3}
\begin{tabular}{lccccccccccccccc}
\toprule
&  \multicolumn{10}{c}{\textbf{en→xx}} & \multicolumn{2}{c}{\textbf{cs→xx}} & \multicolumn{1}{c}{\textbf{ja→xx}} \\
& \textsc{cs} & \textsc{et} & \textsc{ru} & \textsc{sh} & \textsc{uk} & \textsc{is} & \textsc{ar} & \textsc{zh} & \textsc{ja} & \textsc{ko} & \textsc{de} & \textsc{uk} & \textsc{zh} \\
\midrule
\textbf{\small	Baselines} \\
\textsc{\footnotesize	 TOWER-V2 7B } & 71.7 & - & 79.7 & - & - & - & - & 81.9 & - & \textbf{84.1} & 76.8 & - & - \\
\textsc{\footnotesize MadLad400 7B } & 82.7 & 83.2 & 76.8 & - & 82.1 & 71.1 & 72.4 & 73.7 & 81.7 & 78.3 & 81.8 & 82.8 & 76.4\\
\textsc{\footnotesize NLLB 3.3B} & 79.5 & 80.4 & 76.6 & - & 78.3 & 70.1 & 72.7 & 70.3 & 77.9 & 80.3 & 76.9 & 78.9 & 68.4 \\
\cdashlinelr{1-16}
\textsc{\footnotesize Salamandra\textbf{TA}{\footnotesize 2B}} \\
\textsc{\tiny BASE {\tiny + CPT-v1} } & \textcolor{gray}{80.3} & \textcolor{gray}{80.1} & \textcolor{gray}{76.0} & \textcolor{gray}{-} & \textcolor{gray}{69.6} & \textcolor{gray}{-} & \textcolor{gray}{-} & \textcolor{gray}{-} & \textcolor{gray}{-} & \textcolor{gray}{-} & \textcolor{gray}{80.1} & \textcolor{gray}{57.0} & \textcolor{gray}{-} \\
\quad \textsc{\scriptsize + Instruct-v1} & 80.7 & 80.3 & 76.5 & - & 78.0 & - & - & - & - & - & 76.0 & 78.0 & - \\
\quad \quad \textsc{\scriptsize + TRR} & 84.3 & 86.0 & 80.5 & - & 83.3 & - & - & - & - & - & 80.4 & 81.8 & - \\
\quad \quad \textsc{\scriptsize + MBR} & 85.6 & 87.0 & 81.4 & - & 84.0 & - & - & - & - & - & 81.5 & 83.5 & - \\
\cdashlinelr{1-16}
\textsc{\footnotesize Salamandra\textbf{TA}{\footnotesize 7B}} \\
\textsc{\tiny BASE {\tiny + CPT-v1} } & \textcolor{gray}{81.9} & \textcolor{gray}{79.8} & \textcolor{gray}{76.6} & \textcolor{gray}{-} & \textcolor{gray}{78.0} & \textcolor{gray}{-} & \textcolor{gray}{-} & \textcolor{gray}{-} & \textcolor{gray}{-} & \textcolor{gray}{-} & \textcolor{gray}{81.5} & \textcolor{gray}{82.2} & \textcolor{gray}{-} \\
\quad \textsc{\scriptsize + Instruct-v1} & 85.3 & 86.6 & 80.3 & - & 83.8 & - & - & - & - & - & 81.6 & 83.4 & - \\
\quad \quad \textsc{\scriptsize + TRR} & 85.9 & 87.6 & 82.0 & - & 85.0 & - & - & - & - & - & 81.3 & 84.0 & - \\
\quad \quad \textsc{\scriptsize + MBR} & \textbf{87.2} & \textbf{88.7} & \textbf{82.9} & - & 85.9 & - & - & - & - & - & \textbf{82.6 }& \textbf{85.1} & - \\
\cdashlinelr{1-16}
\textsc{\footnotesize Salamandra\textbf{TA}}{\scriptsize \textsc{-v2}} \\
\textsc{\scriptsize Base {\tiny + CPT-v1} {\tiny + CPT-v2} } & \textcolor{gray}{81.1} & \textcolor{gray}{79.3} & \textcolor{gray}{76.2} & \textcolor{gray}{79.4} & \textcolor{gray}{77.0} & \textcolor{gray}{69.3} & \textcolor{gray}{70.6} & \textcolor{gray}{74.7} & \textcolor{gray}{75.5} & \textcolor{gray}{75.9} & \textcolor{gray}{81.5} & \textcolor{gray}{82.5} & \textcolor{gray}{77.3} \\
\quad \textsc{\scriptsize + Instruct-v2} & 83.1 & 85.3 & 79.3 & 83.9 & 84.1 & 77.4 & 71.3 & 81.1 & 80.9 & 80.2 & 80.4 & 82.3 & 77.8 \\
\quad \quad \textsc{\scriptsize + TRR} & 85.3 & 87.3 & 81.8 & 84.9 & 85.1 & 79.7 & 74.2 & 82.7 & 83.3 & 82.5 & 81.3 & 84.2 & 79.6 \\
\quad \quad \textsc{\scriptsize + MBR} & 86.6 & 88.5 & 82.4 & 86.3 & \textbf{86.1} & \textbf{80.7} & \textbf{75.5} & \textbf{83.4} & \textbf{84.1} & 83.6 & 82.5 & \textbf{85.1} & \textbf{80.4} \\
\bottomrule
\end{tabular}
\caption{\textsc{comet} scores on the \wmtplusplus\ test set, comparing our \salamandraTA\ models against several strong baselines. We show the performance at each stage of our method: from the continually pre-trained base models (scores in gray), to the instruction-tuned models, and finally with the application of quality-aware decoding strategies (TRR and MBR). Using Minimum Bayes Risk (MBR) decoding consistently yields the best results.}\label{tab:comet_main_results}
\end{table*}

\subsection{Continual pre-training}

To train the \salamandraTA\ models, we first compile a parallel corpus from publicly available data sources. A comprehensive list of these sources and the corresponding language pairs can be found in Table \ref{tab:data_sources_cp1}. We build two separate training sets: \salamandraTAcpone\ and \salamandraTAcptwo. All data undergo initial filtering using \textsc{LaBSE} \cite{feng-etal-2022-language}, and off-target translations are excluded using the Lingua\footnote{\url{https://github.com/pemistahl/lingua-py}} library. After filtering, the data is de-duplicated and punctuation is normalized with the \texttt{Bifixer} library \cite{ramirez-sanchez-etal-2020-bifixer}. The final corpora are formatted using the prompt template provided in Appendix Figure \ref{fig:translation-templates}. Additional dataset details are available in Appendix \ref{app:dataset}.

Using \salamandraTAcpone\ we continue pre-training  \salamandra\ 2B and 7B with the causal language modeling objective resulting in \salamandraTAtwobase\ and \salamandraTAsevenbase\ models. Then, we use \salamandraTAcptwo\ to continue pre-training \salamandraTAsevenbase. \\

\noindent
\coloredsquare{cp1_color} \salamandraTAcpone: The first corpus, is employed during the initial round of continual pre-training (CPT), with the objective of enhancing the machine translation capabilities of \salamandra\ across European languages. The final dataset has 38 languages across 6.57B sentence pairs and 172 machine translation directions in total pivoting in English, Spanish and Catalan, totaling in 424B tokens. We show in Figure \ref{fig:data_distribution} the data distribution of the \salamandraTAcpone\ corpus.\\

\noindent
\coloredsquare{cp2_color} \salamandraTAcptwo: The second corpus, is used in the subsequent CPT round, where the focus shifts toward expanding coverage to include the additional language pairs featured in the WMT 2025 shared task. It includes 0.39B sentences across 14 languages and 15 directions, amounting to 27B tokens. To avoid the risk of catastrophic forgetting, we subsample 20M sentences for directions already present in \salamandraTAcpone\ ( \textsc{\small en→cs}, \textsc{\small en→et}, \textsc{\small en→ru}, \textsc{\small en→uk}).
For \textsc{\small en→sh} (English-to-Serbian, Latin script), we combined two sources from \salamandraTAcpone{}: English–Serbian (Latin script) data and English–Serbian (Cyrillic script) data, the latter converted to Latin script using rule-based transliteration.
The per-direction data distribution is also shown in Figure~\ref{fig:data_distribution}. Note that we include the English-to-Hindi direction, which is not part of this year’s shared task, in order to support better transfer for related languages such as Bhojpuri.

\subsection{Instruction tuning}

For instruction tuning we build two separate corpus: \salamandraTAitone\ and \salamandraTAittwo. The first, \salamandraTAitone, is used to fine-tune \salamandraTAtwobase\ and \salamandraTAsevenbase\ models into instruction-following models. The second corpus, is used to instruct \salamandraTAsevenbase\ after continue pre-training with \salamandraTAcptwo\ corpus. We format each instruction using the \texttt{chatml} template \cite{chatml}.\\

\noindent
\coloredsquare{cp1_color} \salamandraTAitone: Following prior work on supervised fine-tuning for machine translation \cite{alves2024tower, rei-etal-2024-tower, rei2025towerbridginggeneralitytranslation}, we organize the instruction examples into three categories: pre-translation, translation, and post-translation tasks. The selection of tasks is motivated by the ablation results discussed in Section \ref{sec:results}. The final corpus consists of 135k instructions, with the majority sourced from the \towerblocks\ collection \cite{alves2024tower}. For translation related-tasks we focus on sentence, paragraph and document level data, primarily sourced from \europarl\ \cite{koehn-2005-europarl}. A big part of the data is drawn from multi-parallel datasets such as \flores\ \cite{nllbteam2022language} or \ntrex\ \cite{federmann-etal-2022-ntrex}, where a single source sentence has multiple translations in different target languages. When building the instruction tuning dataset, a naive strategy is to pivot trough different bridge languages across all languages including the complete dataset (e.g. for a given Catalan sentence that aligns to parallel sentences in, Spanish, French, and German, we might generate \textsc{\small ca→es}, \textsc{\small ca→fr}, \textsc{\small ca→de} and \textsc{\small es→ca}, \textsc{\small fr→ca}, \textsc{\small de→ca}). In our dataset we pivoted in five bridge languages: English, Catalan, Spanish, Basque and Galician across all the supported languages. However, this increases the number of duplicate training examples that share identical content on the target or source side. We found that doing this encourages target-side collapse, where the model produces off-target translations because many-to-one alignments blur the mapping between specific source inputs and their intended target languages. To mitigate this, we randomly sampled approximately equal numbers of translation instructions for each language pair. Further details on \salamandraTAitone\ are provided in Appendix \ref{app:dataset}.\\

\noindent
\coloredsquare{cp2_color} \salamandraTAittwo:  The second corpus, consisting of approximately 51k instructions, is constructed to focus on paragraph-level translation, context-aware machine translation, and sentence-level translation for the language directions included in the WMT 2025 shared task. 
To construct paragraph level data we source from \flores-dev, \ntrex\ and \newscommentary\ datasets.
Similar to \salamandraTAitone, we applied random sampling when using multi-parallel datasets. In addition, we included data from \towerblocks\ that we considered relevant to our tasks.  
More details about \salamandraTAittwo\ can be found in Appendix \ref{app:dataset}.

\begin{table*}[!ht]
\centering
\renewcommand{\arraystretch}{1.2}
\begin{tabular}{lcccccc}
\toprule
 & \multicolumn{3}{c}{\textbf{en→xx}} & \multicolumn{3}{c}{\textbf{xx→en}} \\
 & \textsc{comet} & \textsc{metricX} & \textsc{bleu} & \textsc{comet} & \textsc{metricX} & \textsc{bleu} \\
\midrule
\textsc{Salamandra\textbf{TA}{\normalsize 7B}} \textsc{\tiny BASE {\tiny + CPT-v1} } & 0.85 & 1.73 & 34.60 & \textbf{0.88} & 1.15 & 44.22 \\
\midrule
\multicolumn{5}{l}{\textbf{Supervised Finetuning}} \\
\quad MT  & \textbf{0.87} & 1.33 &  \textbf{36.71} & \textbf{0.88} & 1.17 & \textbf{45.02} \\
\quad \quad + Pre-MT + Post-MT & \textbf{0.87} & \textbf{1.14} & 36.42 & \textbf{0.88} & \textbf{1.09} & 45.00 \\
\quad \quad  \quad   + Chat + Code & \textbf{0.87} & 1.36 & 35.58 & \textbf{0.88} & 1.16 & 44.81 \\

\cdashlinelr{1-7}

\quad MT + Post-MT & \textbf{0.87} & 1.33 & 36.57 & \textbf{0.88} & 1.15 & 44.88 \\
\quad MT + Pre-MT  & \textbf{0.87} & 1.33 & 36.34 & \textbf{0.88} & 1.16 & 44.67 \\

\bottomrule
\end{tabular}
\caption{Ablation study on the impact of different supervised fine-tuning tasks for the \textsc{Salamandra\textbf{TA}{\normalsize 7B-Base}} model. We report \textsc{comet}, \textsc{metricX}, and \textsc{bleu} scores for English-to-Other (en→xx) and Other-to-English (xx→en) directions.}\label{tab:ablation_tasks}
\end{table*}

\section{SalamandraTA Models}

The \salamandraTA\ family is composed of two base models, 2B and 7B parameters, which were continually pre-trained on the \salamandraTAcpone\ corpus and subsequently instruction-tuned on \salamandraTAitone. For our submission to the WMT25 General Translation Shared Task, we further adapted the 7B model, resulting in \salamandraTAversiontwo.

\subsection{Adding WMT languages: {\scriptsize \textsc{Salamandra\textbf{TA}}{\footnotesize \textsc{-v2}}} }

To expand the language coverage of \salamandraTA\ and accommodate the additional languages required by the WMT25 General Translation Shared Task, we implemented vocabulary adaptation. We trained a new tokenizer on a corpus comprising the original languages augmented with monolingual text for the new languages not included in the original \salamandra\ tokenizer: Chinese, Korean, Japanese, Arabic, and Bhojpuri.

The old tokenizer was replaced with the new one, which required re-initializing the embedding and unembedding layers. To address this, we modified these layers to ensure that tokens common to both the old and new tokenizers retained their original embeddings. The embeddings for the remaining, newly introduced tokens were initialized as the average of all existing embeddings. We expected this strategy to be particularly successful given that the two tokenizers share over 58\% of their vocabulary. Figure \ref{fig:fertility-barplot} shows the fertility per language pair, comparing our new \salamandra\ tokenizer against previous tokenizer, \textsc{\footnotesize MadLad400} and \textsc{\footnotesize NLLB}. On average, \salamandra\ achieves a fertility of 1.88, outperforming both \textsc{\footnotesize NLLB} (2.00) and \textsc{\footnotesize MadLad400} (2.33) on WMT25 language pairs.

The subsequent section details the continual pre-training stage of our model. This stage aims not only to enhance the model's translation capabilities but also to recover the embeddings of these newly initialized tokens. More details can be found in Appendix \ref{app:tokenizer}.

\subsection{Model training} 

\subsubsection{Continual pre-training}
For this phase, we chose \salamandra-2B and \salamandra-7B as base models, using checkpoints preceding the annealing phase described in \citet{gonzalezagirre2025salamandratechnicalreport}. This choice was intentional: the annealing phase narrows the data sources to shape the model into a general-purpose downstream performer, which we considered misaligned with (or even counterproductive to) our goal of improving translation capabilities.
The training strategy followed a schedule similar to that of the annealing phase. The learning rate was linearly warmed up over the first 2,000 steps, reaching a peak of 3e-5, and then decayed using a cosine schedule down to 3e-6. To mitigate the risk of exploding gradients, we applied gradient clipping with a maximum norm of 1.0 after the warm-up stage.
We used NVIDIA NeMo as the training framework, and all other training hyperparameters were kept consistent with those used in the original \salamandra\ pre-training (see Appendix \ref{app:training} for more details). We trained the 7B model for 105k steps and the 2B model for 50k steps on the \salamandraTAcpone\ corpus tokenized with the original \salamandra\ tokenizer (see Appendix Figure \ref{fig:train_loss_cpt_v1}).

After vocabulary adaptation, we continually pre-train the resulting \salamandraTA-7B model using \salamandraTAcptwo. The training strategy followed the same training configuration as previously described.

\subsubsection{Supervised Fine-tuning}

We fine-tune \salamandraTA\ base models using \texttt{FastChat} framework \cite{zheng2023judging}. Hyperparameter details are provided in Appendix Table \ref{tab:supervised_training-hparams}.

\subsection{Evaluation}

\paragraph{Metrics} We assess translation quality using several metrics. For reference-based evaluation, we report scores from the learned metrics \comet\ \cite{rei-etal-2022-comet}, \textsc{Bleurt} \cite{sellam-etal-2020-bleurt}, and \textsc{MetricX} \cite{juraska-etal-2023-metricx}. For reference-free quality estimation (QE), we use \cometkiwi\ \cite{rei-etal-2022-cometkiwi}, and \textsc{MetricX-QE}. We also report two lexical-based metrics: \textsc{CHRF} \cite{popovic-2015-chrf} and \textsc{Bleu} \cite{papineni-etal-2002-bleu}.

\paragraph{Datasets} We used the \flores-devtest dataset for ablation studies on the \salamandraTA\ models. For evaluating translation quality on the WMT 2025 directions, we primarily relied on the \wmtplusplus\ dataset \cite{deutsch2025wmt24expandinglanguagecoverage}. An exception is the English to Bhojpuri direction, which is not included in \wmtplusplus; for this case, we used \flores-devtest for evaluation.

\paragraph{Baselines} We compare the different \salamandraTA\ variants against the translation LLM \textsc{\footnotesize TOWER-V2 7B} \cite{rei-etal-2024-tower}, as well as dedicated MMNMT models such as \textsc{\footnotesize MadLad400 7B} \cite{10.5555/3666122.3669062} and \textsc{\footnotesize NLLB 3.3B} \cite{nllbteam2022language}.

\paragraph{Decoding strategies} For inference with the baseline, base, and instruction-tuned models, we employ beam search with a beam size of 5. Additionally, we experiment with two alternative decoding approaches: we use diverse beam search \cite{vijayakumar2018diversebeamsearchdecoding}, which promotes output diversity by penalizing similar beams, and two post-decoding strategies applied to the generated candidates: Tuned Re-ranking Decoding (TRR) and Minimum Bayes Risk Decoding (MBR) \cite{eikema-aziz-2020-map} using the \texttt{mbrs} library \cite{deguchi-etal-2024-mbrs}. For diverse beam search we set a beam size of 20 and 5 beam groups. For post-decoding methods, we use \textsc{Comet{\scriptsize-22}} as the quality metric for MBR and \cometkiwi\ for TRR.

\begin{figure*}[t]
\centering
  \includegraphics[width=\linewidth]{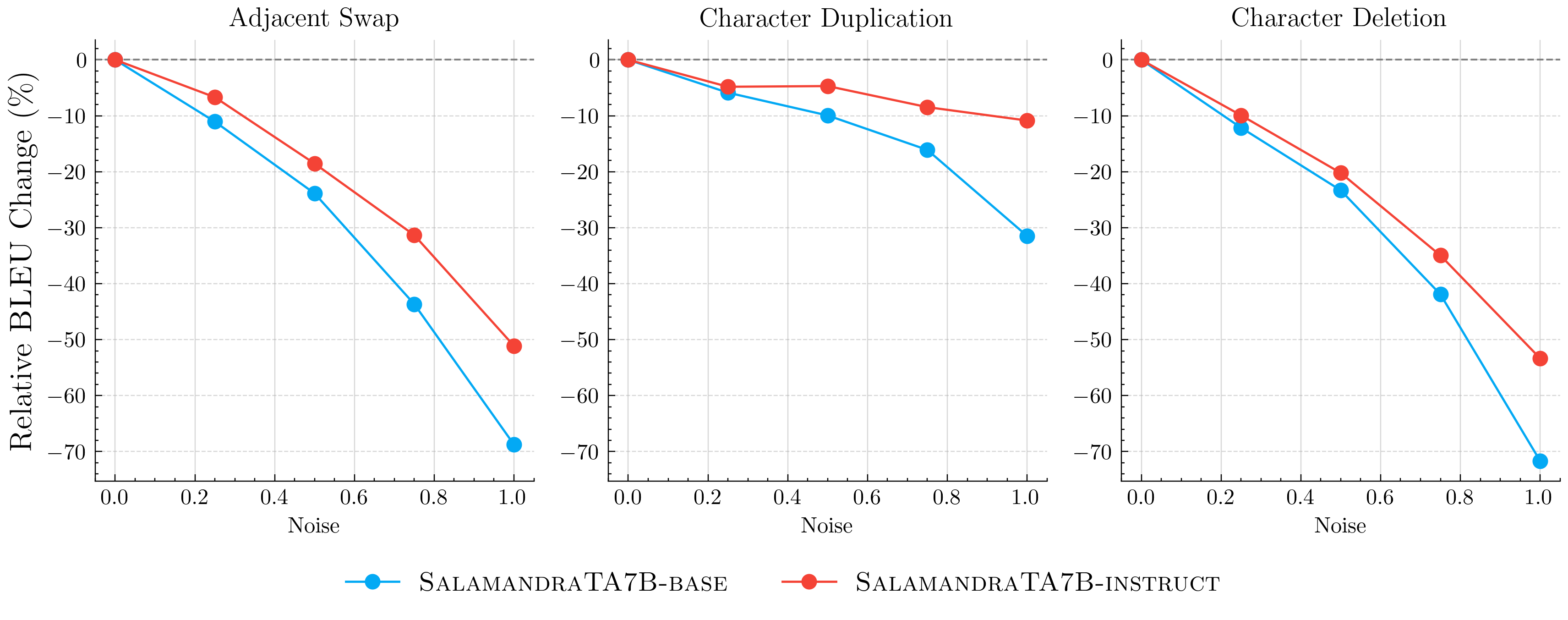}
  \caption{Relative change in \textsc{BLEU} scores (\%) under increasing levels of input noise for three types of character-level perturbations: Adjacent Swap, Character Duplication, and Character Deletion.}\label{fig:relative_bleu_change}
\end{figure*}

\section{Results}\label{sec:results} 

Table \ref{tab:comet_main_results} presents the main translation quality results on the \wmtplusplus\ test set, measured in \comet\ scores for the language directions in the general MT task. We report extra metrics in Appendix \ref{app:extra_metrics}. We additionally evaluate \salamandraTA-2B and \salamandraTA-7B using \comet\ and \textsc{MetricX} for the language directions present in the multilingual subtask and report them in Appendix Table \ref{app:table_multilingual}.

As shown in Table \ref{tab:comet_main_results}, instruction tuning yields significant gains over the CPT baselines, improving the \salamandraTA-7B, \salamandraTA-2B, and \salamandraTAversiontwo\ models by an average of 3.51, 4.40, and 3.60 \comet\ points, respectively. 

Although further adapting the \salamandraTA-7B model to WMT-2025 language pairs initially causes an average performance drop of 1.09 \comet\ points on the language directions shared between \salamandraTA-7B and \salamandraTAversiontwo, this gap is largely mitigated when employing quality-aware decoding strategies. Applying Minimum Bayes Risk (MBR) and Tuned Re-ranking (TRR) decoding strategies reduces this drop to 0.16 and 0.20 \comet\ points, respectively.

\paragraph{On the impact of adding non-MT-Tasks} To better understand the impact of different instruction types on translation quality, we conduct an ablation study of instruction fine-tuning across four main task categories: machine translation (MT), pre-translation tasks (Pre-MT) (e.g., Named Entity Recognition), post-translation tasks (Post-MT) (e.g., Gender Bias Mitigation), and chat/code-related tasks\footnote{This last group includes \towerblocks\ synthetic chat data and code instruction data.}. Table~\ref{tab:ablation_tasks} presents the model's performance after fine-tuning on each of these categories.

Instruction fine-tuning using MT tasks consistently yields the best overall performance across most evaluation metrics, with the exception of \textsc{MetricX}. For \textsc{MetricX}, a combination of MT, Pre-MT, and Post-MT instructions results in slightly improved performance. In contrast, adding only Pre-MT or Post-MT instructions shows no significant difference compared to the MT-only baseline. Incorporating Chat and Code instructions, however, leads to a consistent drop in BLEU scores without measurable gains in other metrics.

Based on these findings, we concluded that for \salamandraTA-2B and 7B, incorporating both Pre-MT and Post-MT tasks alongside MT tasks provided a slight benefit or at least no degradation in performance, leading to their inclusion in the \salamandraTAitone\ dataset. However, for \salamandraTAversiontwo\, which was specifically tailored for the WMT25 General Translation Shared Task, we made a deliberate choice to focus exclusively on MT instructions. While Pre-MT and Post-MT tasks might offer benefits, gathering high-quality, task-specific instruction data for the unique language pairs and domains present in WMT25 would have required significant additional effort beyond the scope of this work.

\paragraph{On the robustness to character noise} Following \citet{peters-martins-2025-translation}, we investigate model robustness by injecting character-level noise into the source sentences of \flores-devtest for the English to Spanish direction using adjacent swaps, duplications, and deletions at different noise levels. Figure~\ref{fig:relative_bleu_change} shows the relative degradation in \textsc{BLEU} score compared to zero-noise baseline. The \salamandraTA\ 7B instruction-tuned model consistently shows greater resilience than the base model across all perturbation types. At the maximum noise level (1.0), the performance degradation of the instruction-tuned model is smaller by 17.63 p.p. for swaps, 20.61 p.p. for duplications, and 18.33 p.p. for deletions. These results demonstrate that instruction tuning effectively improves a model's robustness to character-level input corruptions.

\paragraph{Adding a low-resource language: The case of Bhojpuri} Table~\ref{tab:bhojpuri_results} presents our ablation experiments for English to Bhojpuri translation direction. We find that during CPT, removing the \textsc{en→hi} parallel data causes performance to drop from 9.32 to 0.35 \textsc{BLEU} and from 35.43 to 9.83 \textsc{ChrF}. This result provides clear evidence that the model relies on cross-lingual transfer from Hindi for translating to Bhojpuri. Finally, supervised fine-tuning (\textsc{IT-v2}) improves performance, improving the scores to 11.67 \textsc{BLEU} and 37.75 \textsc{ChrF}. This result shows the effectiveness of fine-tuning on high-quality data in the final stage, even for low-resource language pairs.

\begin{table}[!h]
\centering
\begin{tabular}{lrr}
\toprule
 & \textsc{BLEU} & \textsc{ChrF} \\
\midrule
\textbf{\scriptsize Continual pre-training} & & \\
\textsc{\normalsize \textsc{CPT-v2}} & 9.32 & 35.43  \\
\textsc{\normalsize \textsc{CPT-v2}} (no \textsc{\small en→hi}) & 0.35 & 9.83 \\
\cdashlinelr{1-3}
\textbf{\scriptsize Supervised Finetuning} & & \\
\textsc{\normalsize \textsc{CPT-v2} + \textsc{IT-v2}  } & 11.67 & 37.75 \\

\bottomrule
\end{tabular}
\caption{Ablation results for English$\rightarrow$Bhojpuri translation in terms of \bleu\ and \textsc{ChrF} on \flores\-devtest. The table compares the impact of removing the \textsc{en→hi} direction from the CPT data and the effect of supervised fine-tuning ( \salamandraTAittwo\ ).}
\label{tab:bhojpuri_results}
\end{table}

\section{Submission}

For our WMT25 general and multilingual MT tasks submissions, we apply a chunking strategy, splitting each input instance at \texttt{\textbackslash n\textbackslash n} delimiter prior to translation. We made two submissions using two quality-aware decoding strategies:  Minimum Bayes Risk Decoding employing \comet\ and Tuned Re-ranking relying on \cometkiwi.

\section{Conclusion}

In this paper, we introduced the \salamandraTA\ family of models, a series of powerful, translation LLMs in 2B and 7B scales. Our approach combines a multi-stage training recipe, beginning with continual pre-training on parallel data that pivots through three languages: English, Spanish, and Catalan. This is followed by an instruction tuning stage to align the models with human translation outputs. For our WMT25 submission, we adapted our 7B model to new, non-European languages through vocabulary adaptation and a further round of continual pre-training and supervised fine-tuning.

Our experimental results show that instruction tuning is a critical step which not only improves translation quality but also the model's robustness against character-level noise. Furthermore, our analysis of the English-to-Bhojpuri direction validates the importance of including related languages during pre-training to enable cross-lingual transfer to low-resource pairs.

While our work successfully specializes models for translation and translation-related tasks, we observed that incorporating Chat and Code instructions during the supervised fine-tuning stage leads to a significant drop in translation quality as measured by \bleu. Future work could explore methods to mitigate this trade-off to train machine translation models that can follow general instructions without compromising their specialized translation capabilities.

\section{Acknowledgements}

This work has been promoted and financed by the
Generalitat de Catalunya through the Aina Project.\\

\noindent
This work has been supported by the Spanish
project PID2021-123988OB-C33 funded by MCIN/AEI/10.13039/501100011033/FEDER, UE.\\

\noindent
This work is funded by the Ministerio para la Transformación Digital y de la Función Pública - Funded by EU – NextGenerationEU within the framework of ILENIA Project with reference 2022/TL22/00215337.\\

\noindent
This work is funded by the Ministerio para la Transformación Digital y de la Función Pública and Plan de Recuperación, Transformación y Resiliencia - Funded by EU – NextGenerationEU within the framework of the project Desarrollo Modelos ALIA.

\bibliography{custom}

\appendix

\section{CPT Template}\label{app:cpt_template}

This section presents the template used to prepare parallel data for continued pre-training. 
We used only one single template.
Placeholders:
\begin{itemize}
    \item \{ source \}: source sentence
    \item \{ target \}: target sentence
    \item \{ source\_lang \}: source language name
    \item \{ target\_lang \}: target language name
\end{itemize}

\begin{figure}[h]
\centering
\begin{tcolorbox}[colback=gray!5, colframe=black, title=Template used for CPT, fonttitle=\bfseries, arc=4mm, boxrule=1pt, width=0.4\textwidth]
\begin{Verbatim}[fontsize=\small]

{ source_lang }: { source }
{ target_lang }: { target }
\end{Verbatim}
\end{tcolorbox}
\caption{Template used to format parallel data for CPT.}
\label{fig:translation-templates}
\end{figure}

\section{Prompt templates used to construct translation instructions}\label{app:it_template}

All templates used to construct instructions were adapted from \towerblocks\ \citep{alves2024tower}. Figure~\ref{fig:instruction-templates} shows an example of a template used for translation instructions in our \salamandraTAitone\ and \salamandraTAittwo\ datasets.

\begin{figure*}[h]
\centering
\begin{tcolorbox}[colback=gray!5, colframe=black, title=Template used for IT, fonttitle=\bfseries, arc=4mm, boxrule=1pt, width=0.9\textwidth]
\begin{Verbatim}[fontsize=\small]

Translate the following text from { source_lang } to { target_lang }: 
{ source_lang }: { source }
{ target_lang }: { target } 
\end{Verbatim}
\end{tcolorbox}
\caption{Example of a prompt template used to construct translation instructions for \salamandraTAitone\ and \salamandraTAittwo.}
\label{fig:instruction-templates}
\end{figure*}

\section{Dataset} \label{app:dataset}
\subsection{Continual pre-training v1} \label{app:cpt-v1}

The pre-training corpus for \salamandraTAcpone\ consists of 424 billion tokens of Catalan-centric, Spanish-centric, and English-centric parallel data, including all of the official European languages plus Catalan, Basque, Galician, Asturian, Aragonese and Aranese. It amounts to 6,574,251,526 parallel sentence pairs.

This highly multilingual corpus is predominantly composed of data sourced from OPUS \citep{tiedemann-2012-parallel}, with additional data taken from the NTEU Project \citep{garcia-martinez-etal-2021-neural}, Aina Project,\footnote{\url{https://projecteaina.cat/}} and other sources (see Table~\ref{tab:data_sources_cp1}, and Table~\ref{tab:lang_code_map} shows the mapping between the BCP-47 language code and the language name). Where little parallel Catalan~$\leftrightarrow$~xx data could be found, synthetic Catalan data was generated from the Spanish side of the collected Spanish~$\leftrightarrow$~xx corpora using Projecte Aina’s Spanish-Catalan model.\footnote{\url{https://huggingface.co/projecte-aina/aina-translator-es-ca}} The final distribution of languages is shown in Figure~\ref{fig:data_distribution}.

Datasets with "-BSC" in their names (e.g., BOUA-SYNTH-BSC, DOGV-SYNTH-BSC) are synthetic datasets obtained by machine translating pre-existing monolingual corpora with our own seq-to-seq models. These datasets were generated internally for model training and are not published.

\begin{table}[h!]
\centering
\begin{tabular}{ll}
\toprule
\textbf{Language Code} & \textbf{Language} \\
\midrule
ar & Arabic \\
arn & Aranese \\
ast & Asturian \\
arg & Aragonese \\
bho & Bhojpuri \\
bg & Bulgarian \\
ca & Catalan \\
cs & Czech \\
cy & Welsh \\
da & Danish \\
de & German \\
el & Greek \\
es & Spanish \\
en & English \\
et & Estonian \\
eu & Basque \\
fi & Finnish \\
fr & French \\
ga & Irish \\
gl & Galician \\
hi & Hindi \\
hr & Croatian \\
hu & Hungarian \\
is & Icelandic \\
it & Italian \\
ja & Japanese \\
ko & Korean \\
lt & Lithuanian \\
lv & Latvian \\
mt & Maltese \\
nl & Dutch \\
nn & Norwegian Nynorsk \\
no & Norwegian \\
oc & Occitan \\
pl & Polish \\
pt & Portuguese \\
ro & Romanian \\
ru & Russian \\
sh & Serbian (Latin) \\
sk & Slovak \\
sl & Slovenian \\
sr & Serbian (Cyrillic) \\
sv & Swedish \\
uk & Ukrainian \\
val & Catalan-Valencian \\
zh & Chinese \\
\bottomrule
\end{tabular}
\caption{Mapping from BCP-47 language codes to full language names.}
\label{tab:lang_code_map}
\end{table}
\begin{table*}[htbp]
\centering
\tiny
\begin{tabular}{p{7cm}p{2.2cm}p{2.2cm}p{2.2cm}}
\toprule
\textbf{Dataset} & \textbf{Ca-xx Languages} & \textbf{Es-xx Languages} & \textbf{En-xx Languages} \\
\midrule
AINA \citep{caen_parallel_corpus_2024} & en & & \\ 

\cdashlinelr{1-4}

ARANESE-SYNTH-CORPUS-BSC & arn & & \\ 

\cdashlinelr{1-4}

BOUA-SYNTH-BSC & & val & \\ 

\cdashlinelr{1-4}

BOUMH \citep{pilar2024} & & val & \\ 

\cdashlinelr{1-4}

BOUA-PILAR \citep{pilar2024} & & val & \\ 

\cdashlinelr{1-4}

CCMatrix \citep{schwenk-etal-2021-ccmatrix} & eu & & ga \\ 

\cdashlinelr{1-4}

DGT \citep{steinberger-etal-2012-dgt} & bg, cs, da, de, el, et, fi, fr, ga, hr, hu, lt, lv, mt, nl, pl, pt, ro, sk, sl, sv & da, et, ga, hr, hu, lt, lv, mt, sh, sl & \\ 

\cdashlinelr{1-4}

DOGV-SYNTH-BSC & & val & \\ 

\cdashlinelr{1-4}

DOGV-PILAR \citep{pilar2024} & & val & \\ 

\cdashlinelr{1-4}
ELRC-EMEA \citep{elrc_emea_2020}  & bg, cs, da, hu, lt, lv, mt, pl, ro, sk, sl & et, hr, lv, ro, sk, sl & \\ 

\cdashlinelr{1-4}

EMEA \citep{tiedemann-2012-parallel} & bg, cs, da, el, fi, hu, lt, mt, nl, pl, ro, sk, sl, sv & et, mt & \\ 

\cdashlinelr{1-4}

EUBookshop \citep{skadins-etal-2014-billions} & lt, pl, pt & cs, da, de, el, fi, fr, ga, it, lv, mt, nl, pl, pt, ro, sk, sl, sv & cy, ga \\ 

\cdashlinelr{1-4}

Europarl \citep{koehn-2005-europarl} & & bg, cs, da, el, en, fi, fr, hu, lt, lv, nl, pl, pt, ro, sk, sl, sv & \\ 

\cdashlinelr{1-4}

Europat \citep{heafield-etal-2022-europat} & & en, hr & no \\

\cdashlinelr{1-4}

GAITU Corpus \citep{caeu_parallel_corpus_2024} & & & eu \\ 

\cdashlinelr{1-4}

KDE4 \citep{tiedemann-2012-parallel} & bg, cs, da, de, el, et, eu, fi, fr, ga, gl, hr, it, lt, lv, nl, pl, pt, ro, sk, sl, sv & bg, ga, hr & cy, ga, nn, oc

 \\ 

\cdashlinelr{1-4}

GlobalVoices \citep{CASMACAT2018, tiedemann-2012-parallel} & bg, de, fr, it, nl, pl, pt & bg, de, fr, pt & \\ 

\cdashlinelr{1-4}

GNOME \citep{gnome,tiedemann-2012-parallel} & eu, fr, ga, gl, pt & ga & cy, ga, nn \\ 

\cdashlinelr{1-4}

JRC-Arquis \citep{steinberger-etal-2006-jrc} & cs, da, et, fr, lt, lv, mt, nl, pl, ro, sv & & et \\ 

\cdashlinelr{1-4}

LES-CORTS-VALENCIANES-SYNTH-BSC & & val & \\ 

\cdashlinelr{1-4}

MaCoCu \citep{non-etal-2022-macocu} & en & & hr, mt, uk \\ 

\cdashlinelr{1-4}

MultiCCAligned \citep{elkishky_ccaligned_2020} & bg, cs, de, el, et, fi, fr, hr, hu, it, lt, lv, nl, pl, ro, sk, sv & bg, fi, fr, hr, it, lv, nl, pt & bg, cy, da, et, fi, hr, hu, lt, lv, no, sl, sr, uk \\ 

\cdashlinelr{1-4}

MultiHPLT \citep{de-gibert-etal-2024-new} & en, et, fi, ga, hr, mt & fi, ga, gl, hr, mt, nn, sr \\ 

\cdashlinelr{1-4}

MultiParaCrawl \citep{banon-etal-2020-paracrawl} & bg, da & de, en, fr, ga, hr, hu, it, mt, pt & bg, cs, da, de, el, et, fi, fr, ga, hr, hu, lt, lv, mt, nn, pl, ro, sk, sl, uk \\ 

\cdashlinelr{1-4}

MultiUN \citep{eisele-chen-2010-multiun} & & fr & \\ 

\cdashlinelr{1-4}

News-Commentary \citep{tiedemann-2012-parallel} & & fr & \\ 

\cdashlinelr{1-4}

NLLB \citep{nllbteam2022language} & bg, da, el, en, et, fi, fr, gl, hu, it, lt, lv, pt, ro, sk, sl & bg, cs, da, de, el, et, fi, fr, hu, it, lt, lv, nl, pl, pt, ro, sk, sl, sv & bg, cs, cy, da, de, el, et, fi, fr, ga, hr, hu, it, lt, lv, mt, nl, no, oc, pl, pt, ro, ru, sk, sl, sr, sv, uk \\

\cdashlinelr{1-4}

NÓS Authentic Corpus \citep{nos_authentic_corpus_2023} & & & gl \\ 

\cdashlinelr{1-4}

NÓS Synthetic Corpus \citep{nos_en_gl_synth_corpus_2023} & & & gl \\

\cdashlinelr{1-4}

NTEU \citep{garcia-martinez-etal-2021-neural} & bg, cs, da, de, el, en, et, fi, fr, ga, hr, hu, it, lt, lv, mt, nl, pl, pt, ro, sk, sl, sv & da, et, ga, hr, lt, lv, mt, ro, sk, sl, sv & \\

\cdashlinelr{1-4}

OpenSubtitles \citep{lison-tiedemann-2016-opensubtitles2016} & bg, cs, da, de, el, et, eu, fi, gl, hr, hu, lt, lv, nl, pl, pt, ro, sk, sl, sv & da, de, fi, fr, hr, hu, it, lv, nl & bg, cs, de, el, et, hr, fi, fr, hr, hu, no, sl, sr\\ 

\cdashlinelr{1-4}

OPUS-100 \citep{zhang-etal-2020-improving,tiedemann-2012-parallel} & en & & gl \\ 

\cdashlinelr{1-4}

StanfordNLP-NMT \cite{luong-manning-2016-achieving,luong-etal-2015-effective,luong-manning-2015-stanford}
 & & & cs \\ 

\cdashlinelr{1-4}

Tatoeba \citep{tiedemann-2012-parallel} & de, pt & pt & \\ 

\cdashlinelr{1-4}

TildeModel \citep{rozis-skadins-2017-tilde} & bg & et, hr, lt, lv, mt & \\ 

\cdashlinelr{1-4}

UNPC \citep{ziemski-etal-2016-united} & & en, fr & ru \\ 

\cdashlinelr{1-4}

PILAR-VALENCIAN-AUTH \citep{pilar2024} & & val & \\ 

\cdashlinelr{1-4}

PILAR-VALENCIAN-SYNTH \citep{pilar2024} & & val & \\ 

\cdashlinelr{1-4}

WikiMatrix \citep{schwenk-etal-2021-wikimatrix} & bg, cs, da, de, el, et, eu, fi, fr, gl, hr, hu, it, lt, nl, pl, pt, ro, sk, sl, sv & bg, en, fr, hr, it, pt & oc, sh \\ 

\cdashlinelr{1-4}

Wikimedia & & & cy, nn \\ 

\cdashlinelr{1-4}

XLENT \citep{el-kishky-etal-2021-xlent} & eu, ga, gl & ga & cy, et, ga, gl, hr, oc, sh \\ 
\bottomrule

\end{tabular}
\caption{Data sources of \textbf{ \textsc{CPT-v1}}.}
\label{tab:data_sources_cp1}
\end{table*}

\subsection{Continual pre-training v2}
\label{app:cpt-v2}

In \salamandraTAcptwo\, we focused on the language pairs featured in the WMT 2025 shared task.  
For pairs involving European languages, we reused part of the data from \salamandraTAcpone.  
Specifically, we sampled 20M sentence pairs each for English–Czech, English–Estonian, and English–Russian from the \salamandraTAcpone\ data. For English–Serbian (Latin), we included the authentic English–Serbian (Latin) parallel dataset from \salamandraTAcpone. Additionally, we transliterated the Serbian side of the English–Serbian (Cyrillic) dataset into Latin script, taking advantage of the one-to-one correspondence between the two scripts.  
For English–Icelandic, Czech–Ukrainian, and Czech–German, we used the WMT 2025 Translation Task Training Data.\footnote{\url{https://www2.statmt.org/wmt25/mtdata/}}

For language pairs involving non-European languages, we used sentence-level data from the WMT 2025 Translation Task Training Data.
The Chinese side of all datasets were first processed using the Hanzi Identifier to detect Traditional Chinese,%
\footnote{\url{https://github.com/tsroten/hanzidentifier}}
which was subsequently converted to Simplified Chinese using OpenCC.%
\footnote{\url{https://github.com/BYVoid/OpenCC}}
We also included paragraph-level English–Arabic data by concatenating sentences from \newscommentary.

We created two versions of \salamandraTAcptwo. The first included only the language pairs featured in the WMT25 shared task.  
In the second, we additionally included English–Hindi data from the OPUS corpora CCMatrix \citep{schwenk-etal-2021-ccmatrix}, MultiHPLT \citep{de-gibert-etal-2024-new}, NLLB \citep{nllbteam2022language}, and Samanantar \citep{ramesh-etal-2022-samanantar}, to support the model’s performance on Bhojpuri (which uses the Devanagari script).

The pre-training corpus for \salamandraTAcptwo\ wihout English-Hindi consists of 24 billion tokens, amounting to 366,179,935 parallel sentence pairs.  
For \salamandraTAcptwo\ with English-Hindi, the corpus contains 26 billion tokens and 393,507,678 parallel sentence pairs. The data distribution is shown in Figure~\ref{fig:data_distribution}, and the corresponding sources are listed in Table~\ref{tab:data_sources_cp2}.

\begin{table*}[h]
\centering
\begin{tabular}{ll}
\toprule
\textbf{Source} & \textbf{Language Pair} \\
\midrule

WMT 2025 Translation Task Training Data                  & en-ar \\
                                                         & en-zh \\
                                                         & cs-de \\
                                                         & en-ko \\
                                                         & en-ja \\
                                                         & ja-zh \\
                                                         & en-is \\
                                                         & cs-uk \\
                                                         & en-bho \\
\cdashlinelr{1-2}                                                         
\newscommentary\ (paragraph-level)                                         & en-ar \\
\cdashlinelr{1-2}
\textsc{CCMatrix} \citep{schwenk-etal-2021-ccmatrix}                                                 & en-hi \\
\cdashlinelr{1-2}
\textsc{MultiHPLT} \citep{de-gibert-etal-2024-new}                                               & en-hi \\
\cdashlinelr{1-2}
\textsc{NLLB}  \citep{nllbteam2022language}                                                   & en-hi \\
\cdashlinelr{1-2}
\textsc{Samanantar}  \citep{ramesh-etal-2022-samanantar}                                             & en-hi \\
\cdashlinelr{1-2}
\salamandraTAcpone                & en-cs \\
                                                         & en-et \\
                                                         & en-ru \\
                                                         & en-uk \\
                                                         & en-sh \\
\bottomrule
\end{tabular}
\caption{Data sources of \textbf{ \textsc{CPT-v2}}.}
\label{tab:data_sources_cp2}
\end{table*}

As shown in Section~\ref{sec:results}, continual pre-training with Hindi data led to better performance, particularly for Bhojpuri.

\subsection{Instruction tuning v1} \label{app:it_v1_data}

During \salamandraTAitone\, the model was fine-tuned on \textasciitilde135k instructions, primarily targeting machine translation performance for Catalan, English, and Spanish. Additional instruction data for other European and closely related Iberian languages was also included.

\begin{figure}[t]
\centering
  \includegraphics[width=\linewidth]{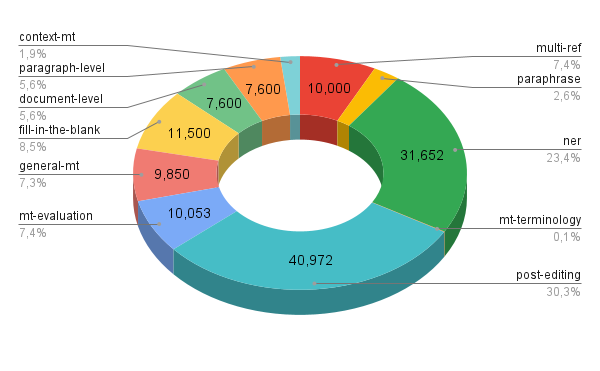}
  \caption {Distribution of tasks in \salamandraTAitone\ .}\label{fig:task_distribution_instruct_v1}
\end{figure}

A portion of our fine-tuning data comes directly from, or is sampled from \towerblocks. 
While tasks related to machine translation are included, it is important to note that no chat data was used in the fine-tuning process. The final distribution of tasks is shown in Figure~\ref{fig:task_distribution_instruct_v1}. The full list of tasks included in \salamandraTAitone\ is shown in Table~\ref{tab:task_overview_instruct_v1}.

\begin{table*}[htbp]
\centering
\footnotesize
\begin{tabular}{p{2.3cm} p{2cm} p{5.4cm} p{3.4cm} r}
\toprule
\textbf{Category} & \textbf{Task} & \textbf{Source} & \textbf{Languages} & \textbf{Count} \\
\midrule

Pre-Translation & Named-entity & \textsc{AnCora-Ca-NER} & ca & 12,059 \\
& Recognition & \textsc{BasqueGLUE}, \textsc{EusIE} & eu & 4,304 \\
& & \textsc{SLI NERC} Galician Gold Corpus & gl & 6,483 \\
& & \textsc{TowerBlocks}: {\tiny \textsc{MultiCoNER} 2022 and 2023 Dev} & pt & 854 \\
& & \textsc{TowerBlocks}: {\tiny \textsc{MultiCoNER} 2022 and 2023 Dev} & nl & 800 \\
& & \textsc{TowerBlocks}: {\tiny \textsc{MultiCoNER} 2022 and 2023 Dev} & es & 1,654 \\
& & \textsc{TowerBlocks}: {\tiny \textsc{MultiCoNER} 2022 and 2023 Dev} & en & 1,671 \\
& & \textsc{TowerBlocks}: {\tiny \textsc{MultiCoNER} 2022 and 2023 Dev} & ru & 800 \\
& & \textsc{TowerBlocks}: {\tiny \textsc{MultiCoNER} 2022 and 2023 Dev} & it & 858 \\
& & \textsc{TowerBlocks}: {\tiny \textsc{MultiCoNER} 2022 and 2023 Dev} & fr & 857 \\
& & \textsc{TowerBlocks}: {\tiny \textsc{MultiCoNER} 2022 and 2023 Dev} & de & 1,312 \\

\cdashlinelr{1-5}

Translation & Multi-reference Translation & \textsc{TowerBlocks}: {\tiny \textsc{Tatoeba} Dev} & mixed & 10,000 \\
\cdashlinelr{2-5}
& Terminology-aware & \textsc{\textsc{TowerBlocks}}: {\tiny \textsc{WMT21 Terminology Dev} } & en-ru & 50 \\
& Translation & \textsc{TowerBlocks}: {\tiny \textsc{WMT21 Terminology Dev}} & en-fr & 29 \\
\cdashlinelr{2-5}
& Fill-in-the-Blank & Non-public & Five pivot languages (ca, es, eu, gl, en) paired with European languages (cs, da, de, el, et, fi, fr, ga, hr, hu, it, lt, lv, mt, nl, pl, pt, ro, sk, sl, sv) & 11,500 \\
\cdashlinelr{2-5}
& General Machine Translation & \textsc{TowerBlocks}: {\tiny \textsc{WMT14} to \textsc{WMT21}, \textsc{NTREX}, \textsc{Flores Dev}, \textsc{FRMT}, \textsc{QT21}, \textsc{ApeQuest}, \textsc{OPUS} (Quality Filtered), \textsc{MT-GenEval} } & nl-en, en-ru, it-en, fr-en, es-en, en-fr, ru-en, fr-de, en-nl, de-fr & 500 \\
&  & \textsc{Flores Dev}, \textsc{NTREX} & Four pivot languages (es, ca, eu, gl) paired with the rest of languages. We sample 50 instances for each pair. & 9350 \\
\cdashlinelr{2-5}
& Document-level Translation & Non-public & Two pivot languages (es, en) paired with European languages (bg, cs, da, de, el, et, fi, fr, hu, it, lt, lv, nl, pl, pt, ro, ru, sk, sv) & 7,600 \\
\cdashlinelr{2-5}
& Paragraph-level Translation & Non-public & Two pivot languages (es, en) paired with European languages (bg, cs, da, de, el, et, fi, fr, hu, it, lt, lv, nl, pl, pt, ro, ru, sk, sv) & 7,600 \\
\cdashlinelr{2-5}
& Context-Aware & \textsc{TowerBlocks}: {\tiny \textsc{MT-GenEval}} & en-it & 348 \\
& Translation & & en-ru & 454 \\
& & & en-fr & 369 \\
& & & en-nl & 417 \\
& & & en-es & 431 \\
& & & en-de & 558 \\

\cdashlinelr{1-5}
Post-Translation & Paraphrase & \textsc{TowerBlocks}: {\tiny \textsc{PAWS-X Dev}} & mixed & 3,521 \\
\cdashlinelr{2-5}
& Machine Translation Evaluation & \textsc{TowerBlocks} (sample): {\tiny \textsc{WMT20} to \textsc{WMT22 Metrics MQM}, \textsc{WMT17} to \textsc{WMT22 Metrics Direct Assessments} } & en-ru, en-pl, ru-en, en-de, en-ru, de-fr, de-en, en-de & 353 \\

& & Non-public & Four pivot languages (eu, es, ca, gl) paired with European languages (bg, cs, da, de, el, en, et, fi, fr, ga, hr, hu, it, lt, lv, mt, nl, pl, pt, ro, sk, sl, sv) & 9,700 \\

\cdashlinelr{2-5}
& Automatic Post & \textsc{TowerBlocks}: {\tiny \textsc{QT21, ApeQuest}} & en-fr & 6,133 \\
& Editing & \textsc{TowerBlocks}: {\tiny \textsc{QT21, ApeQuest}} & en-nl & 9,077 \\
& & \textsc{TowerBlocks}: {\tiny \textsc{QT21, ApeQuest}} & en-pt & 5,762 \\
& & \textsc{TowerBlocks}: {\tiny \textsc{QT21, ApeQuest}} & de-en & 10,000 \\
& & \textsc{TowerBlocks}: {\tiny \textsc{QT21, ApeQuest}} & en-de & 10,000 \\

\midrule
\textbf{Total} & & & & \textbf{135,404} \\
\bottomrule
\end{tabular}
\caption{Overview of tasks, data sources, language coverage, and counts in \textbf{\textsc{IT-v1}}.}
\label{tab:task_overview_instruct_v1}
\end{table*}

\subsection{Instruction tuning v2} \label{app:it_v2_data}

\begin{figure}[t]
\centering
  \includegraphics[width=\linewidth]{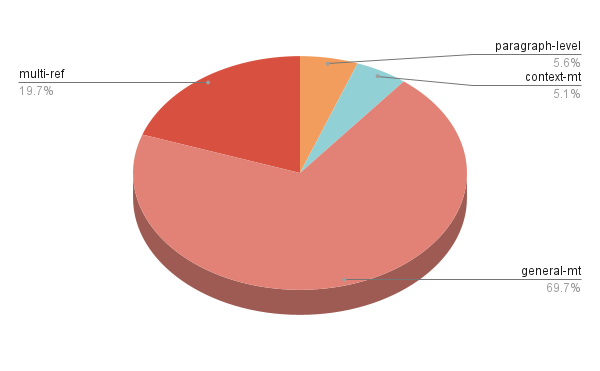}
  \caption {Distribution of tasks in \salamandraTAittwo\ .}\label{fig:task_distribution_instruct_v2}
\end{figure}

In \salamandraTAittwo\, we focused on the languages pairs featured in the WMT 2025 shared task. We included paragraph-level data during instruction tuning to support paragraph-level translation.
We constructed this data by concatenating adjacent sentences (randomly grouping 2, 3, or 4) from the same article or document in \flores-dev, \ntrex, and \newscommentary.
To prevent over-representation of these sources, we sampled approximately equal amounts of paragraph-level data for each language pair.
Serbian Cyrillic data from \flores-dev was transliterated into Serbian Latin. 
In addition, we included data from \towerblocks\ that we considered relevant to our tasks. The instruction tuning dataset is summarized in Table~\ref{tab:task_overview_instruct_v2} and the distribution of tasks is shown in Figure~\ref{fig:task_distribution_instruct_v2}.

\begin{table*}[t]
\centering
\footnotesize
\begin{tabular}{p{2.3cm} p{3.7cm} p{4.2cm} p{2.5cm} r}
\toprule
\textbf{Category} & \textbf{Task} & \textbf{Source} & \textbf{Languages} & \textbf{Count} \\
\midrule

Translation & Paragraph-level Translation & \textsc{Flores Dev} & en-ar & 30 \\
\multicolumn{3}{c}{} & en-bho & 30 \\
\multicolumn{3}{c}{} & en-ja & 30 \\
\multicolumn{3}{c}{} & en-uk & 30 \\
\multicolumn{3}{c}{} & en-ru & 21 \\
\multicolumn{3}{c}{} & cs-uk & 30 \\
\multicolumn{3}{c}{} & ja-zh & 30 \\
\multicolumn{3}{c}{} & en-zh & 30 \\
\multicolumn{3}{c}{} & en-ko & 30 \\
\multicolumn{3}{c}{} & en-et & 30 \\
\multicolumn{3}{c}{} & en-is & 30 \\
\multicolumn{3}{c}{} & en-sh & 30 \\
\multicolumn{3}{c}{} & en-cs & 30 \\
\multicolumn{3}{c}{} & cs-de & 30 \\

& & \ntrex & en-ja & 58 \\
\multicolumn{3}{c}{} & en-uk & 58 \\
\multicolumn{3}{c}{} & en-ru & 50 \\
\multicolumn{3}{c}{} & cs-uk & 58 \\
\multicolumn{3}{c}{} & ja-zh & 58 \\
\multicolumn{3}{c}{} & en-zh & 58 \\
\multicolumn{3}{c}{} & en-ko & 58 \\
\multicolumn{3}{c}{} & en-et & 58 \\
\multicolumn{3}{c}{} & en-is & 58 \\
\multicolumn{3}{c}{} & en-sh & 58 \\
\multicolumn{3}{c}{} & en-cs & 58 \\
\multicolumn{3}{c}{} & cs-de & 58 \\

& & \textsc{News Commentary} & en-zh & 250 \\
\multicolumn{3}{c}{} & cs-de & 250 \\
\multicolumn{3}{c}{} & en-cs & 250 \\
\multicolumn{3}{c}{} & en-de & 250 \\
\multicolumn{3}{c}{} & en-ja & 250 \\
\multicolumn{3}{c}{} & ja-zh & 250 \\
\multicolumn{3}{c}{} & en-ru & 250 \\

\cdashlinelr{2-5}

& Context-Aware Translation & \textsc{TowerBlocks}: {\tiny \textsc{MT-GenEval}} & en-it & 348 \\
\multicolumn{3}{c}{} & en-fr & 369 \\
\multicolumn{3}{c}{} & en-nl & 417 \\
\multicolumn{3}{c}{} & en-es & 431 \\
\multicolumn{3}{c}{} & en-de & 558 \\
\multicolumn{3}{c}{} & en-ru & 454 \\

\cdashlinelr{2-5}

& Multi-reference Translation & \textsc{TowerBlocks}: {\tiny \textsc{Tatoeba} Dev} & mixed & 10,000 \\

\cdashlinelr{2-5}

& General Machine Translation & \textsc{TowerBlocks}: {\tiny \textsc{WMT14} to \textsc{WMT21}, \textsc{NTREX}, \textsc{Flores Dev}, \textsc{FRMT}, \textsc{QT21}, \textsc{ApeQuest}, \textsc{OPUS} (Quality Filtered), \textsc{MT-GenEval}} & en-ru & 22,112 \\
\multicolumn{3}{c}{} & en-zh & 10,521 \\
\multicolumn{3}{c}{} & en-ko & 2,782 \\

\midrule
\textbf{Total} & & & & \textbf{50,841} \\
\bottomrule
\end{tabular}
\caption{Overview of tasks, data sources, language coverage, and counts in \textbf{\textsc{IT-v2}}.}
\label{tab:task_overview_instruct_v2}
\end{table*}

\section{Tokenizer}\label{app:tokenizer}

\begin{figure*}[t]
    \centering
    \includegraphics[width=\textwidth]{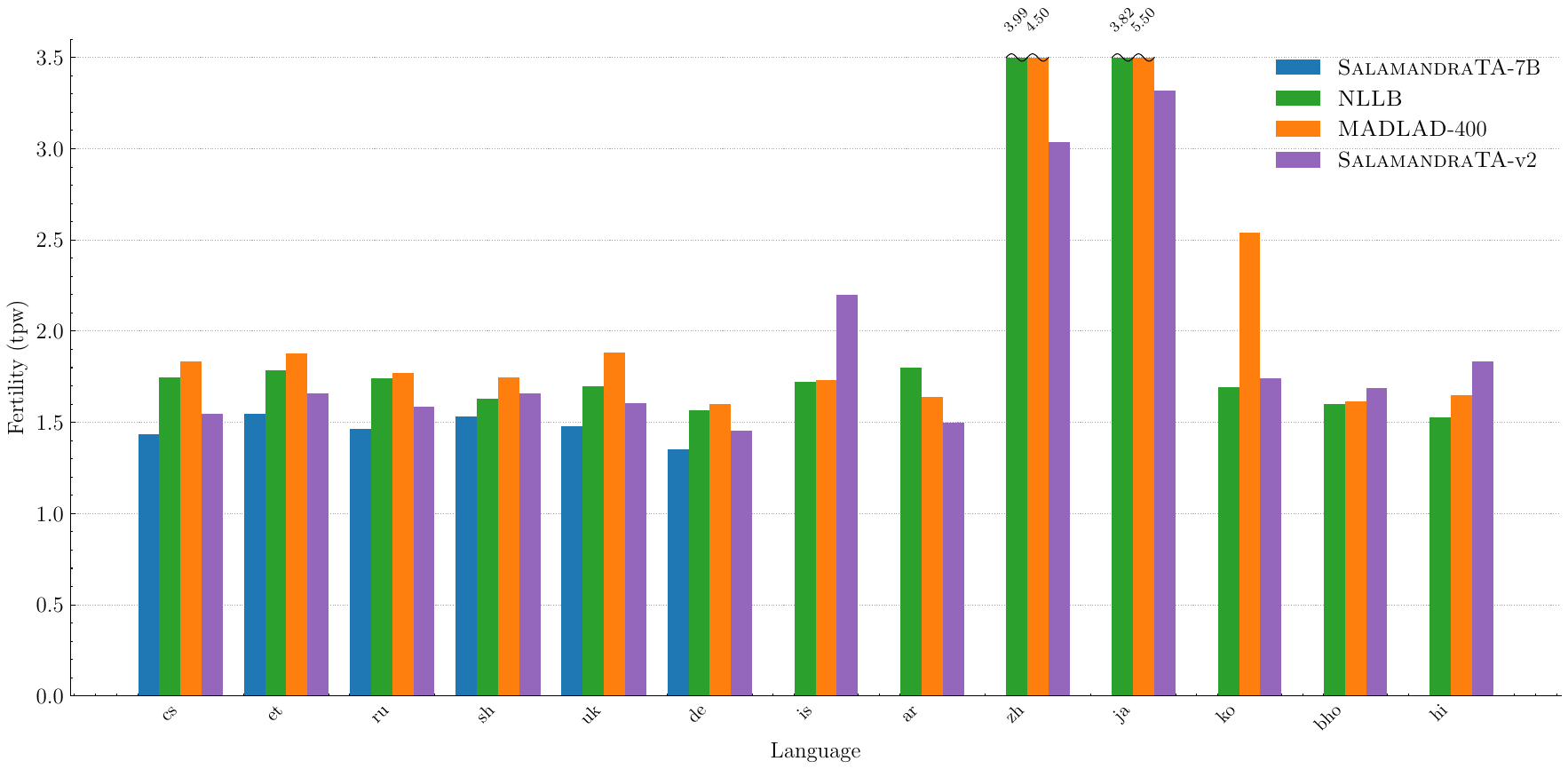}
    \caption{ Tokenization fertility comparison across 13 languages from the \textsc{FLORES-200} dataset. Fertility is shown on the vertical axis for each language on the horizontal axis. Results are presented for four multilingual models: \textsc{SalamandraTA-7B}, \textsc{NLLB}, \textsc{MADLAD-400}, and \textsc{SalamandraTA-v2}. }
    \label{fig:fertility-barplot}
\end{figure*}

We evaluated the trained tokenizer using fertility metric on the \flores\ dataset (see Figure~\ref{fig:fertility-barplot}). For a given tokenizer $T$ and a set of sentences $S$, fertility is defined as the ratio of the total number of tokens produced by $T$ to the total number of words in $S$. Formally:

\begin{equation}
\text{Fertility}(T, S) = \frac{\# \text{tokens in } T(S)}{\# \text{words in } S}
\end{equation}

The results in Figure~\ref{fig:fertility-barplot} indicate that \textsc{Salamandra\textbf{TA}{\normalsize 7B}-v2} consistently achieves the lowest fertility scores on average among WMT25 languages.

\section{Training} \label{app:training}

\begin{table}[ht]
\centering
\small
\caption{Hyperparameters for \salamandraTA\ continual pre-training.}
\label{tab:training-hparams}
\begin{tabular}{ll}
\toprule
\textbf{Hyperparameter} & \textbf{Value} \\
\midrule
Micro Batch Size             & 2 \\
Global Batch Size            & 512 \\
Optimizer                    & Distributed Fused Adam \\
Learning Rate                & 3e-5 \\
Minimum LR                  & 3e-6 \\
Weight Decay                 & 0.1 \\
Betas                        & (0.9, 0.95) \\
LR Scheduler                 & CosineAnnealing \\
Warmup Steps                 & 2048 \\
Mixed Precision              & AMP O2 \\
Sequence Length              & 8,192 \\
Gradient Sync DType          & bfloat16 \\
\bottomrule
\end{tabular}
\end{table}

\begin{table}[ht]
\centering
\small
\caption{Hyperparameters for \salamandraTA\ supervised-fine tuning.}
\label{tab:supervised_training-hparams}
\begin{tabular}{ll}
\toprule
\textbf{Hyperparameter} & \textbf{Value} \\
\midrule
Train epochs                          & 1 \\
Train batch size per device              & 1 \\
Gradient accumulation steps              & 16 \\
Learning rate                            & 1e-5 \\
Weight decay                             & 0 \\
Warmup ratio                             & 0.03 \\
LR scheduler                             & Cosine \\
Model max length                         & 8,192 \\
\bottomrule
\end{tabular}
\end{table}

\begin{figure*}[t]
    \centering
    \includegraphics[width=\textwidth]{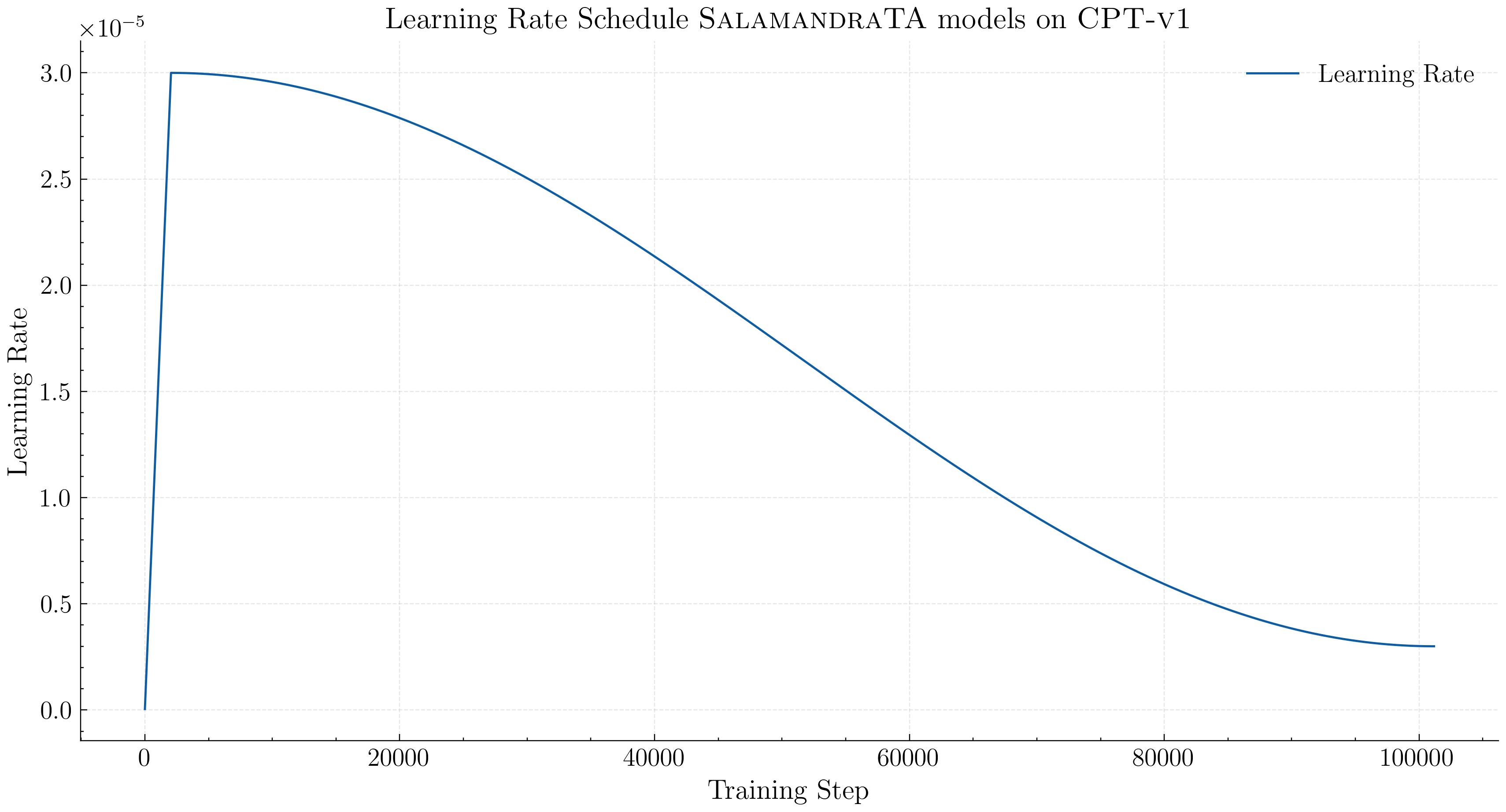}
    \caption{Learning Rate for the \salamandraTA-7B and  \salamandraTA-2B on  \salamandraTAcpone.}
    \label{fig:lr_cpt_v1}
\end{figure*}

\begin{figure*}[t]
    \centering
    \includegraphics[width=\textwidth]{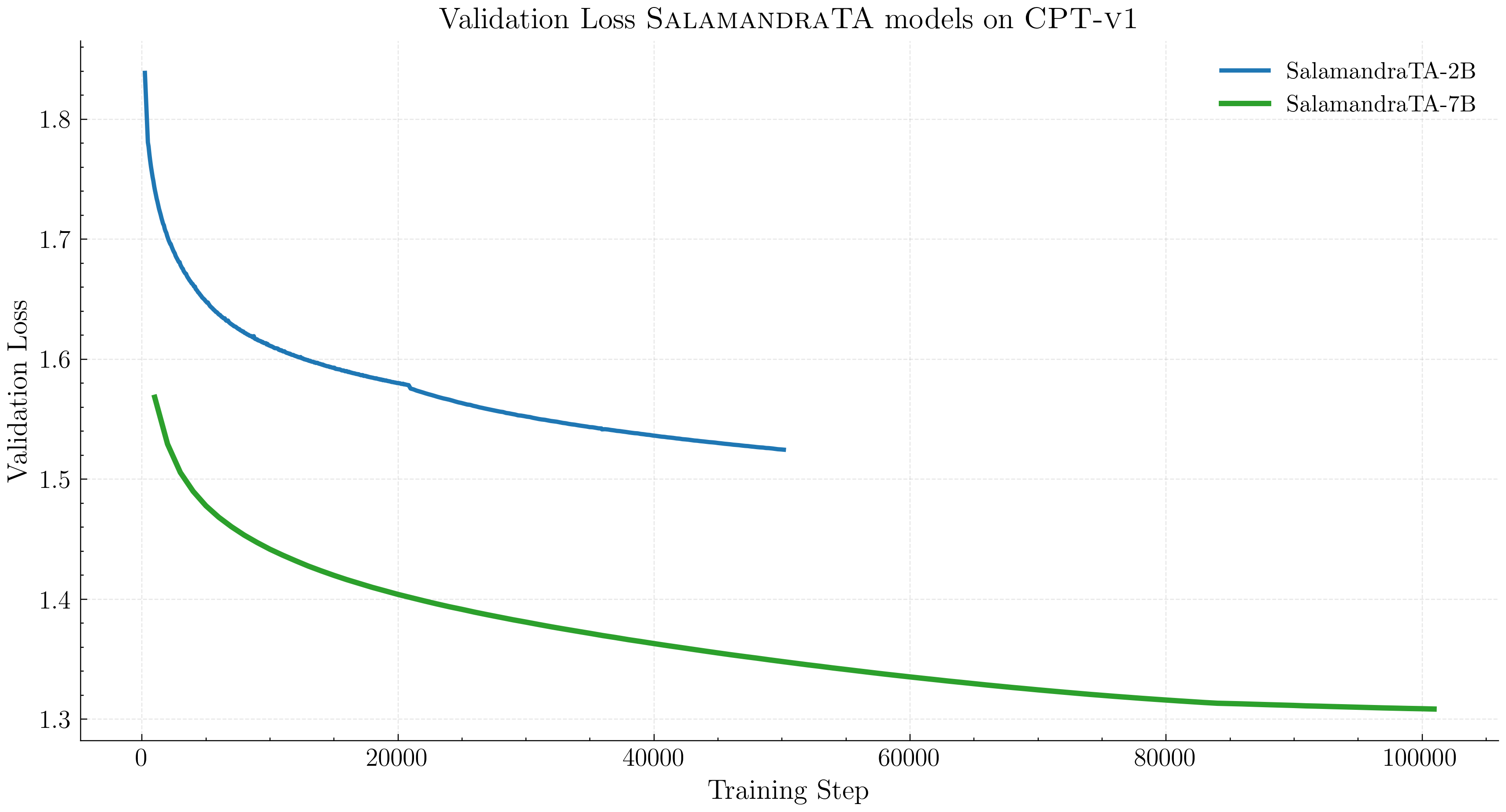}
    \caption{Validation loss for the \salamandraTA-7B and  \salamandraTA-2B on  \salamandraTAcpone.}
    \label{fig:val_loss_cpt_v1}
\end{figure*}

\begin{figure*}[t]
    \centering
    \includegraphics[width=\textwidth]{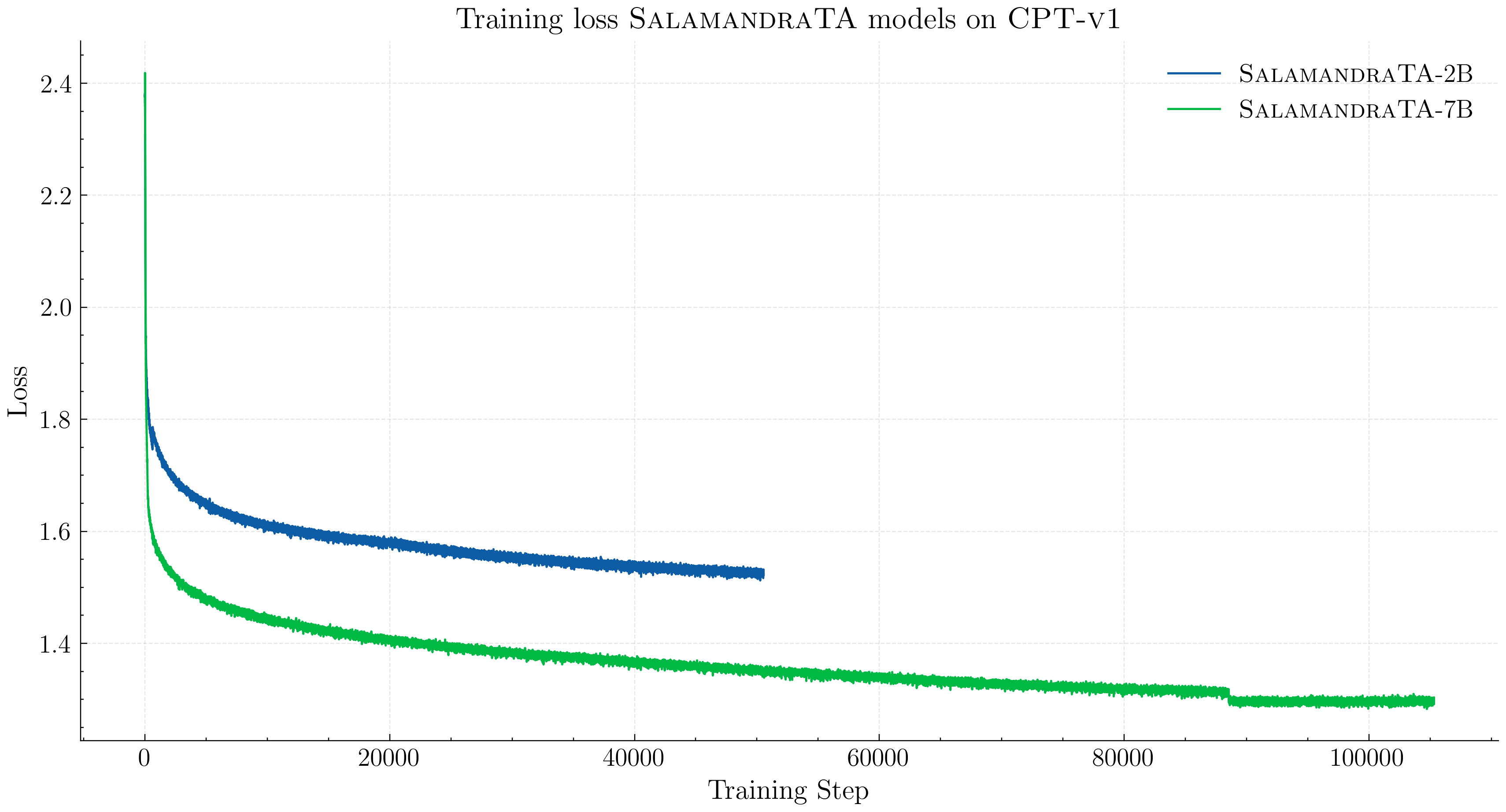}
    \caption{Training loss for the \salamandraTA-7B and  \salamandraTA-2B on  \salamandraTAcpone.}
    \label{fig:train_loss_cpt_v1}
\end{figure*}

\section{Results}\label{app:extra_metrics}

\begin{table*}
\centering
\setlength{\tabcolsep}{4.75pt}
\footnotesize
\renewcommand{\arraystretch}{1.3}
\begin{tabular}{lccccccccccccccc}
\toprule
&  \multicolumn{10}{c}{\textbf{en→xx}} & \multicolumn{2}{c}{\textbf{cs→xx}} & \multicolumn{1}{c}{\textbf{ja→xx}} \\
& \textsc{cs} & \textsc{et} & \textsc{ru} & \textsc{sh} & \textsc{uk} & \textsc{is} & \textsc{ar} & \textsc{zh} & \textsc{ja} & \textsc{ko} & \textsc{de} & \textsc{uk} & \textsc{zh} \\
\midrule
\textbf{\small	Baselines} \\
\textsc{\footnotesize	 TOWER-V2 7B } & 11.1 & - & 21.2 & - & - & - & - & 35.6 & - & 24.7 & 18.4 & - & - \\
\textsc{\footnotesize MadLad400 7B } & 28.4 & 27.2 & 22.5 & - & 26.8 & 17.5 & 6.9 & 30.2 & 19.8 & 25.3 & 25.2 & 20.9 & 20.8 \\
\textsc{\footnotesize NLLB 3.3B} & 23.0 & 21.8 & 20.7 & - & 23.4 & 16.2 & 6.9 & 23.9 & 13.6 & 22.5 & 19.4 & 16.4 & 15.5 \\
\cdashlinelr{1-16}
\textsc{\footnotesize Salamandra\textbf{TA}{\footnotesize 2B}} \\
\textsc{\tiny BASE {\tiny + CPT-v1} } & \textcolor{gray}{17.9} & \textcolor{gray}{19.4} & \textcolor{gray}{18.1} & \textcolor{gray}{-} & \textcolor{gray}{9.5} & \textcolor{gray}{-} & \textcolor{gray}{-} & \textcolor{gray}{-} & \textcolor{gray}{-} & \textcolor{gray}{-} & \textcolor{gray}{19.8} & \textcolor{gray}{3.5} & \textcolor{gray}{-} \\
\quad \textsc{\scriptsize + Instruct-v1} & 17.1 & 12.1 & 13.7 & - & 14.6 & - & - & - & - & - & 10.2 & 10.7 & -
 \\
\quad \quad \textsc{\scriptsize + TRR} & 24.7 & 23.5 & 19.4 & - & 24.9 & - & - & - & - & - & 20.5 & 17.5 & - \\
\quad \quad \textsc{\scriptsize + MBR} & 25.1 & 22.1 & 19.4 & - & 24.6 & - & - & - & - & - & 21.1 & 17.4 & - \\

\cdashlinelr{1-16}
\textsc{\footnotesize Salamandra\textbf{TA}{\footnotesize 7B}} \\
\textsc{\tiny BASE {\tiny + CPT-v1} } & \textcolor{gray}{25.9} & \textcolor{gray}{25.1} & \textcolor{gray}{20.2} & \textcolor{gray}{-} & \textcolor{gray}{25.6} & \textcolor{gray}{-} & \textcolor{gray}{-} & \textcolor{gray}{-} & \textcolor{gray}{-} & \textcolor{gray}{-} & \textcolor{gray}{24.9} & \textcolor{gray}{20.1} & \textcolor{gray}{-} \\
\quad \textsc{\scriptsize + Instruct-v1} & 29.0 & 27.7 & 22.2 & - & 28.7 & - & - & - & - & - & 24.4 & 20.9 & - \\
\quad \quad \textsc{\scriptsize + TRR} & 26.4 & 25.4 & 21.2 & - & 27.1 & - & - & - & - & - & 22.4 & 19.6 & - \\
\quad \quad \textsc{\scriptsize + MBR} & 26.8 & 25.9 & 20.9 & - & 27.1 & - & - & - & - & - & 23.5 & 20.1 & - \\
\cdashlinelr{1-16}
\textsc{\footnotesize Salamandra\textbf{TA}}{\scriptsize \textsc{-v2}} \\
\textsc{\scriptsize Base {\tiny + CPT-v1} {\tiny + CPT-v2} } & \textcolor{gray}{25.6} & \textcolor{gray}{24.7} & \textcolor{gray}{19.6} & \textcolor{gray}{26.1} & \textcolor{gray}{24.1} & \textcolor{gray}{16.9} & \textcolor{gray}{5.3} & \textcolor{gray}{33.0} & \textcolor{gray}{11.9} & \textcolor{gray}{17.4} & \textcolor{gray}{24.8} & \textcolor{gray}{20.6} & \textcolor{gray}{20.1} \\
\quad \textsc{\scriptsize + Instruct-v2} & 27.3 & 25.7 & 19.5 & 27.8 & 29.2 & 17.6 & 6.0 & 36.6 & 14.4 & 18.8 & 20.0 & 19.1 & 22.3 \\
\quad \quad \textsc{\scriptsize + TRR} & 26.5 & 25.1 & 21.0 & 26.6 & 26.7 & 17.4 & 6.1 & 35.8 & 17.7 & 20.9 & 22.6 & 20.3 & 22.3 \\
\quad \quad \textsc{\scriptsize + MBR} & 26.1 & 25.4 & 20.4 & 27.0 & 27.5 & 17.5 & 6.3 & 36.2 & 16.7 & 20.9 & 22.7 & 20.6 & 22.1 \\

\bottomrule
\end{tabular}
\caption{\textsc{Bleu} scores on the \wmtplusplus\ test set, comparing our \salamandraTA\ models against several strong baselines. We show the performance at each stage of our method: from the continually pre-trained base models (scores in gray), to the instruction-tuned models.}
\end{table*}
\begin{table*}
\centering
\setlength{\tabcolsep}{4.75pt}
\footnotesize
\renewcommand{\arraystretch}{1.3}
\begin{tabular}{lccccccccccccccc}
\toprule
&  \multicolumn{10}{c}{\textbf{en→xx}} & \multicolumn{2}{c}{\textbf{cs→xx}} & \multicolumn{1}{c}{\textbf{ja→xx}} \\
& \textsc{cs} & \textsc{et} & \textsc{ru} & \textsc{sh} & \textsc{uk} & \textsc{is} & \textsc{ar} & \textsc{zh} & \textsc{ja} & \textsc{ko} & \textsc{de} & \textsc{uk} & \textsc{zh} \\
\midrule
\textbf{\small	Baselines} \\
\textsc{\footnotesize	 TOWER-V2 7B } & 39.6 & - & 49.7 & - & - & - & - & 32.5 & - & 32.1 & 49.2 & - & - \\
\textsc{\footnotesize MadLad400 7B } & 55.0 & 57.8 & 49.7 & - & 53.2 & 43.4 & 36.2 & 27.7 & 28.0 & 31.5 & 54.7 & 47.8 & 20.6 \\
\textsc{\footnotesize NLLB 3.3B} & 49.7 & 51.7 & 46.6 & - & 48.6 & 40.9 & 35.9 & 22.4 & 23.6 & 29.6 & 47.7 & 42.8 & 15.9 \\
\cdashlinelr{1-16}
\textsc{\footnotesize Salamandra\textbf{TA}{\footnotesize 2B}} \\
\textsc{\tiny BASE {\tiny + CPT-v1} } & \textcolor{gray}{48.4} & \textcolor{gray}{51.7} & \textcolor{gray}{44.5} & \textcolor{gray}{-} & \textcolor{gray}{33.6} & \textcolor{gray}{-} & \textcolor{gray}{-} & \textcolor{gray}{-} & \textcolor{gray}{-} & \textcolor{gray}{-} & \textcolor{gray}{49.7} & \textcolor{gray}{15.5} & \textcolor{gray}{-} \\
\quad \textsc{\scriptsize + Instruct-v1} & 49.3 & 47.6 & 44.9 & - & 45.9 & - & - & - & - & - & 44.7 & 40.4 & - \\
\quad \quad \textsc{\scriptsize + TRR} & 52.7 & 55.7 & 48.6 & - & 52.3 & - & - & - & - & - & 51.4 & 45.5 & - \\
\quad \quad \textsc{\scriptsize + MBR} & 52.5 & 55.0 & 48.4 & - & 51.9 & - & - & - & - & - & 51.8 & 46.0 & - \\

\cdashlinelr{1-16}
\textsc{\footnotesize Salamandra\textbf{TA}{\footnotesize 7B}} \\
\textsc{\tiny BASE {\tiny + CPT-v1} } & \textcolor{gray}{52.8} & \textcolor{gray}{55.6} & \textcolor{gray}{48.7} & \textcolor{gray}{-} & \textcolor{gray}{52.3} & \textcolor{gray}{-} & \textcolor{gray}{-} & \textcolor{gray}{-} & \textcolor{gray}{-} & \textcolor{gray}{-} & \textcolor{gray}{54.0} & \textcolor{gray}{47.9} & \textcolor{gray}{-} \\
\quad \textsc{\scriptsize + Instruct-v1} & 55.9 & 58.4 & 50.7 & - & 55.1 & - & - & - & - & - & 54.4 & 48.6 & - \\
\quad \quad \textsc{\scriptsize + TRR} & 54.0 & 57.3 & 50.1 & - & 54.2 & - & - & - & - & - & 52.9 & 47.8 & - \\
\quad \quad \textsc{\scriptsize + MBR} & 54.4 & 57.2 & 50.1 & - & 54.2 & - & - & - & - & - & 53.9 & 48.2 & - \\
\cdashlinelr{1-16}
\textsc{\footnotesize Salamandra\textbf{TA}}{\scriptsize \textsc{-v2}} \\
\textsc{\scriptsize Base {\tiny + CPT-v1} {\tiny + CPT-v2} } & \textcolor{gray}{52.6} & \textcolor{gray}{54.7} & \textcolor{gray}{48.2} & \textcolor{gray}{54.5} & \textcolor{gray}{51.7} & \textcolor{gray}{42.2} & \textcolor{gray}{34.6} & \textcolor{gray}{28.7} & \textcolor{gray}{22.8} & \textcolor{gray}{26.3} & \textcolor{gray}{54.4} & \textcolor{gray}{47.9} & \textcolor{gray}{22.0} \\
\quad \textsc{\scriptsize + Instruct-v2} & 53.9 & 56.8 & 48.7 & 56.8 & 54.9 & 43.8 & 35.5 & 32.7 & 26.9 & 28.5 & 52.2 & 47.2 & 21.1  \\
\quad \quad \textsc{\scriptsize + TRR} & 54.3 & 57.2 & 50.0 & 56.0 & 54.2 & 44.6 & 36.2 & 32.5 & 28.2 & 28.8 & 53.2 & 48.4 & 21.7  \\
\quad \quad \textsc{\scriptsize + MBR} & 54.0 & 57.3 & 49.6 & 56.5 & 54.4 & 44.4 & 36.3 & 32.8 & 27.9 & 28.9 & 53.7 & 48.4 & 21.6 \\

\bottomrule
\end{tabular}
\caption{\textsc{CHRF} scores on the \wmtplusplus\ test set, comparing our \salamandraTA\ models against several strong baselines. We show the performance at each stage of our method: from the continually pre-trained base models (scores in gray), to the instruction-tuned models. }
\end{table*}

\begin{table*}
\centering
\setlength{\tabcolsep}{4.75pt}
\footnotesize
\renewcommand{\arraystretch}{1.3}
\begin{tabular}{lccccccccccccccc}
\toprule
&  \multicolumn{10}{c}{\textbf{en→xx}} & \multicolumn{2}{c}{\textbf{cs→xx}} & \multicolumn{1}{c}{\textbf{ja→xx}} \\
& \textsc{cs} & \textsc{et} & \textsc{ru} & \textsc{sh} & \textsc{uk} & \textsc{is} & \textsc{ar} & \textsc{zh} & \textsc{ja} & \textsc{ko} & \textsc{de} & \textsc{uk} & \textsc{zh} \\
\midrule
\textbf{\small	Baselines} \\
\textsc{\footnotesize	 TOWER-V2 7B } & 6.69 & - & 4.16 & - & - & - & - & 3.83 & - & 3.70 & 2.25 & - & - \\
\textsc{\footnotesize MadLad400 7B } & 4.28 & 4.14 & 5.50 & - & 4.18 & 7.18 & 7.75 & 6.60 & 4.49 & 5.98 & 1.73 & 4.04 & 6.05 \\
\textsc{\footnotesize NLLB 3.3B} & 5.95 & 6.03 & 6.38 & - & 6.64 & 8.46 & 7.71 & 7.91 & 6.09 & 5.74 & 2.74 & 6.45 & 8.12 \\
\cdashlinelr{1-16}
\textsc{\footnotesize Salamandra\textbf{TA}{\footnotesize 2B}} \\
\textsc{\tiny BASE {\tiny + CPT-v1} } & \textcolor{gray}{5.03} & \textcolor{gray}{5.08} & \textcolor{gray}{5.58} & \textcolor{gray}{-} & \textcolor{gray}{6.53} & \textcolor{gray}{-} & \textcolor{gray}{-} & \textcolor{gray}{-} & \textcolor{gray}{-} & \textcolor{gray}{-} & \textcolor{gray}{2.02} & \textcolor{gray}{6.24} & \textcolor{gray}{-} \\
\quad \textsc{\scriptsize + Instruct-v1} & 3.99 & 4.19 & 4.90 & - & 5.28 & - & - & - & - & - & 2.40 & 5.10 & - \\
\quad \quad \textsc{\scriptsize + TRR} & 3.02 & 2.67 & 3.86 & - & 3.82 & - & - & - & - & - & 1.63 & 3.91 & - \\
\quad \quad \textsc{\scriptsize + MBR} & 3.02 & 2.82 & 3.83 & - & 3.92 & - & - & - & - & - & 1.63 & 3.85 & - \\

\cdashlinelr{1-16}
\textsc{\footnotesize Salamandra\textbf{TA}{\footnotesize 7B}} \\
\textsc{\tiny BASE {\tiny + CPT-v1} } & \textcolor{gray}{4.43} & \textcolor{gray}{5.24} & \textcolor{gray}{5.30} & \textcolor{gray}{-} & \textcolor{gray}{5.58} & \textcolor{gray}{-} & \textcolor{gray}{-} & \textcolor{gray}{-} & \textcolor{gray}{-} & \textcolor{gray}{-} & \textcolor{gray}{1.73} & \textcolor{gray}{4.12} & \textcolor{gray}{-} \\
\quad \textsc{\scriptsize + Instruct-v1} & 2.87 & 2.30 & 3.76 & - & 3.60 & - & - & - & - & - & 1.52 & 3.46 & - \\
\quad \quad \textsc{\scriptsize + TRR} & 2.51 & 2.00 & 3.21 & - & 3.11 & - & - & - & - & - & 1.48 & 3.23 & - \\
\quad \quad \textsc{\scriptsize + MBR} & 2.48 & 1.96 & 3.19 & - & 3.12 & - & - & - & - & - & 1.43 & 3.22 & - \\
\cdashlinelr{1-16}
\textsc{\footnotesize Salamandra\textbf{TA}}{\scriptsize \textsc{-v2}} \\
\textsc{\scriptsize Base {\tiny + CPT-v1} {\tiny + CPT-v2} } & \textcolor{gray}{4.91} & \textcolor{gray}{5.26} & \textcolor{gray}{5.52} & \textcolor{gray}{6.54} & \textcolor{gray}{5.97} & \textcolor{gray}{8.24} & \textcolor{gray}{6.87} & \textcolor{gray}{5.79} & \textcolor{gray}{5.89} & \textcolor{gray}{6.40} & \textcolor{gray}{1.73} & \textcolor{gray}{4.03} & \textcolor{gray}{5.08}  \\
\quad \textsc{\scriptsize + Instruct-v2} & 3.60 & 2.79 & 4.13 & 4.44 & 3.44 & 5.26 & 8.48 & 4.03 & 4.59 & 5.15 & 1.77 & 3.83 & 4.66 \\
\quad \quad \textsc{\scriptsize + TRR} & 2.81 & 2.08 & 3.30 & 3.96 & 2.98 & 4.43 & 7.47 & 3.60 & 3.98 & 4.50 & 1.50 & 3.25 & 4.13  \\
\quad \quad \textsc{\scriptsize + MBR} & 2.79 & 2.12 & 3.35 & 4.00 & 2.99 & 4.57 & 7.73 & 3.62 & 4.00 & 4.51 & 1.49 & 3.28 & 4.21  \\

\bottomrule
\end{tabular}
\caption{\textsc{MetricX} scores on the \wmtplusplus\ test set, comparing our \salamandraTA\ models against several strong baselines. We show the performance at each stage of our method: from the continually pre-trained base models (scores in gray), to the instruction-tuned models.}
\end{table*}
\begin{table*}
\centering
\setlength{\tabcolsep}{4.75pt}
\footnotesize
\renewcommand{\arraystretch}{1.3}
\begin{tabular}{lccccccccccccccc}
\toprule
&  \multicolumn{10}{c}{\textbf{en→xx}} & \multicolumn{2}{c}{\textbf{cs→xx}} & \multicolumn{1}{c}{\textbf{ja→xx}} \\
& \textsc{cs} & \textsc{et} & \textsc{ru} & \textsc{sh} & \textsc{uk} & \textsc{is} & \textsc{ar} & \textsc{zh} & \textsc{ja} & \textsc{ko} & \textsc{de} & \textsc{uk} & \textsc{zh} \\
\midrule
\textbf{\small	Baselines} \\
\textsc{\footnotesize	 TOWER-V2 7B } & 55.9 & - & 61.7 & - & - & - & - & 61.8 & - & 59.8 & 62.4 & - & - \\
\textsc{\footnotesize MadLad400 7B } & 69.3 & 71.2 & 58.7 & - & 62.4 & 52.4 & 39.2 & 53.2 & 53.8 & 52.8 & 68.9 & 61.5 & 54.5 \\
\textsc{\footnotesize NLLB 3.3B} & 65.2 & 67.7 & 58.3 & - & 60.4 & 51.0 & 40.3 & 48.5 & 45.9 & 53.2 & 62.9 & 58.6 & 43.8 \\
\cdashlinelr{1-16}
\textsc{\footnotesize Salamandra\textbf{TA}{\footnotesize 2B}} \\
\textsc{\tiny BASE {\tiny + CPT-v1} } & \textcolor{gray}{66.2} & \textcolor{gray}{67.3} & \textcolor{gray}{57.9} & \textcolor{gray}{-} & \textcolor{gray}{49.8} & \textcolor{gray}{-} & \textcolor{gray}{-} & \textcolor{gray}{-} & \textcolor{gray}{-} & \textcolor{gray}{-} & \textcolor{gray}{67.1} & \textcolor{gray}{29.1} & \textcolor{gray}{-} \\
\quad \textsc{\scriptsize + Instruct-v1} & 68.5 & 69.5 & 59.7 & - & 61.7 & - & - & - & - & - & 66.1 & 59.4 & - \\
\quad \quad \textsc{\scriptsize + TRR} & 70.7 & 73.3 & 62.4 & - & 64.5 & - & - & - & - & - & 68.0 & 61.2 & - \\
\quad \quad \textsc{\scriptsize + MBR} & 70.7 & 73.1 & 61.9 & - & 64.4 & - & - & - & - & - & 68.3 & 62.6 & - \\

\cdashlinelr{1-16}
\textsc{\footnotesize Salamandra\textbf{TA}{\footnotesize 7B}} \\
\textsc{\tiny BASE {\tiny + CPT-v1} } & \textcolor{gray}{68.3} & \textcolor{gray}{67.1} & \textcolor{gray}{58.2} & \textcolor{gray}{-} & \textcolor{gray}{60.5} & \textcolor{gray}{-} & \textcolor{gray}{-} & \textcolor{gray}{-} & \textcolor{gray}{-} & \textcolor{gray}{-} & \textcolor{gray}{68.5} & \textcolor{gray}{63.2} & \textcolor{gray}{-} \\
\quad \textsc{\scriptsize + Instruct-v1} & 72.5 & 75.4 & 62.6 & - & 66.7 & - & - & - & - & - & 69.5 & 64.4 & -  \\
\quad \quad \textsc{\scriptsize + TRR} & 72.6 & 75.7 & 64.2 & - & 67.4 & - & - & - & - & - & 69.2 & 65.0 & - \\
\quad \quad \textsc{\scriptsize + MBR} & 73.1 & 75.9 & 63.9 & - & 67.6 & - & - & - & - & - & 70.0 & 64.9 & - \\
\cdashlinelr{1-16}
\textsc{\footnotesize Salamandra\textbf{TA}}{\scriptsize \textsc{-v2}} \\
\textsc{\scriptsize Base {\tiny + CPT-v1} {\tiny + CPT-v2} } & \textcolor{gray}{67.0} & \textcolor{gray}{66.8} & \textcolor{gray}{58.0} & \textcolor{gray}{68.5} & \textcolor{gray}{59.3} & \textcolor{gray}{52.0} & \textcolor{gray}{41.5} & \textcolor{gray}{54.2} & \textcolor{gray}{48.9} & \textcolor{gray}{50.3} & \textcolor{gray}{68.5} & \textcolor{gray}{63.3} & \textcolor{gray}{55.5} \\
\quad \textsc{\scriptsize + Instruct-v2} & 69.8 & 74.1 & 62.2 & 73.3 & 67.6 & 58.4 & 38.5 & 61.4 & 54.1 & 55.5 & 69.0 & 64.6 & 55.7 \\
\quad \quad \textsc{\scriptsize + TRR} & 71.8 & 75.2 & 63.8 & 73.7 & 67.7 & 59.2 & 39.5 & 62.6 & 55.6 & 56.8 & 69.6 & 65.3 & 57.1 \\
\quad \quad \textsc{\scriptsize + MBR} & 71.8 & 75.4 & 63.8 & 74.1 & 67.8 & 59.2 & 39.2 & 62.4 & 55.6 & 56.8 & 69.5 & 65.4 & 56.8 \\

\bottomrule
\end{tabular}
\caption{\textsc{Bleurt} scores on the \wmtplusplus\ test set, comparing our \salamandraTA\ models against several strong baselines. We show the performance at each stage of our method: from the continually pre-trained base models (scores in gray), to the instruction-tuned models.}
\end{table*}

\begin{table*}
\centering
\setlength{\tabcolsep}{4.75pt}
\footnotesize
\renewcommand{\arraystretch}{1.3}
\begin{tabular}{lccccccccccccccc}
\toprule
&  \multicolumn{10}{c}{\textbf{en→xx}} & \multicolumn{2}{c}{\textbf{cs→xx}} & \multicolumn{1}{c}{\textbf{ja→xx}} \\
& \textsc{cs} & \textsc{et} & \textsc{ru} & \textsc{sh} & \textsc{uk} & \textsc{is} & \textsc{ar} & \textsc{zh} & \textsc{ja} & \textsc{ko} & \textsc{de} & \textsc{uk} & \textsc{zh} \\
\midrule
\textbf{\small	Baselines} \\
\textsc{\footnotesize	 TOWER-V2 7B } & 4.87 & - & 2.28 & - & - & - & - & 2.46 & - & 1.74 & 3.50 & - & - \\
\textsc{\footnotesize MadLad400 7B } & 3.38 & 3.38 & 3.89 & - & 2.94 & 4.95 & 5.31 & 6.00 & 3.50 & 3.66 & 3.28 & 3.31 & 8.32 \\
\textsc{\footnotesize NLLB 3.3B} & 4.83 & 4.92 & 4.80 & - & 4.91 & 6.11 & 4.61 & 7.83 & 4.48 & 3.21 & 6.35 & 5.16 & 9.65 \\
\cdashlinelr{1-16}
\textsc{\footnotesize Salamandra\textbf{TA}{\footnotesize 2B}} \\
\textsc{\tiny BASE {\tiny + CPT-v1} } & \textcolor{gray}{3.73} & \textcolor{gray}{4.02} & \textcolor{gray}{3.46} & \textcolor{gray}{-} & \textcolor{gray}{5.48} & \textcolor{gray}{-} & \textcolor{gray}{-} & \textcolor{gray}{-} & \textcolor{gray}{-} & \textcolor{gray}{-} & \textcolor{gray}{3.97} & \textcolor{gray}{4.46} & \textcolor{gray}{-} \\
\quad \textsc{\scriptsize + Instruct-v1} & 2.89 & 3.46 & 3.21 & - & 3.67 & - & - & - & - & - & 4.12 & 3.38 & -
 \\
\quad \quad \textsc{\scriptsize + TRR} & 1.78 & 1.74 & 1.96 & - & 2.02 & - & - & - & - & - & 2.59 & 1.94 & - \\
\quad \quad \textsc{\scriptsize + MBR} & 1.86 & 1.92 & 2.01 & - & 2.21 & - & - & - & - & - & 2.68 & 2.03 & - \\

\cdashlinelr{1-16}
\textsc{\footnotesize Salamandra\textbf{TA}{\footnotesize 7B}} \\
\textsc{\tiny BASE {\tiny + CPT-v1} } & \textcolor{gray}{3.40} & \textcolor{gray}{4.15} & \textcolor{gray}{3.27} & \textcolor{gray}{-} & \textcolor{gray}{3.71} & \textcolor{gray}{-} & \textcolor{gray}{-} & \textcolor{gray}{-} & \textcolor{gray}{-} & \textcolor{gray}{-} & \textcolor{gray}{3.17} & \textcolor{gray}{2.73} & \textcolor{gray}{-} \\
\quad \textsc{\scriptsize + Instruct-v1} & 1.82 & 1.75 & 2.07 & - & 2.14 & - & - & - & - & - & 2.69 & 1.87 & -  \\
\quad \quad \textsc{\scriptsize + TRR} & 1.49 & 1.42 & 1.63 & - & 1.69 & - & - & - & - & - & 2.46 & 1.58 & - \\
\quad \quad \textsc{\scriptsize + MBR} & 1.52 & 1.44 & 1.74 & - & 1.80 & - & - & - & - & - & 2.46 & 1.60 & - \\
\cdashlinelr{1-16}
\textsc{\footnotesize Salamandra\textbf{TA}}{\scriptsize \textsc{-v2}} \\
\textsc{\scriptsize Base {\tiny + CPT-v1} {\tiny + CPT-v2} } & \textcolor{gray}{3.66} & \textcolor{gray}{4.22} & \textcolor{gray}{3.43} & \textcolor{gray}{4.12} & \textcolor{gray}{4.17} & \textcolor{gray}{5.91} & \textcolor{gray}{3.88} & \textcolor{gray}{3.90} & \textcolor{gray}{3.89} & \textcolor{gray}{3.70} & \textcolor{gray}{3.00} & \textcolor{gray}{2.68} & \textcolor{gray}{4.86}  \\
\quad \textsc{\scriptsize + Instruct-v2} & 2.44 & 2.04 & 2.39 & 2.70 & 2.07 & 3.04 & 5.11 & 2.52 & 2.65 & 2.54 & 3.06 & 2.27 & 4.30  \\
\quad \quad \textsc{\scriptsize + TRR} & 1.67 & 1.40 & 1.67 & 2.26 & 1.66 & 2.35 & 3.95 & 2.14 & 2.07 & 1.93 & 2.50 & 1.57 & 3.75 \\
\quad \quad \textsc{\scriptsize + MBR} & 1.83 & 1.50 & 1.78 & 2.24 & 1.73 & 2.55 & 4.20 & 2.22 & 2.22 & 2.04 & 2.51 & 1.75 & 3.86 \\

\bottomrule
\end{tabular}
\caption{\textsc{MetricX-QE} scores on the \wmtplusplus\ test set, comparing our \salamandraTA\ models against several strong baselines. We show the performance at each stage of our method: from the continually pre-trained base models (scores in gray), to the instruction-tuned models.}
\end{table*}
\begin{table*}
\centering
\setlength{\tabcolsep}{4.75pt}
\footnotesize
\renewcommand{\arraystretch}{1.3}
\begin{tabular}{lccccccccccccccc}
\toprule
&  \multicolumn{10}{c}{\textbf{en→xx}} & \multicolumn{2}{c}{\textbf{cs→xx}} & \multicolumn{1}{c}{\textbf{ja→xx}} \\
& \textsc{cs} & \textsc{et} & \textsc{ru} & \textsc{sh} & \textsc{uk} & \textsc{is} & \textsc{ar} & \textsc{zh} & \textsc{ja} & \textsc{ko} & \textsc{de} & \textsc{uk} & \textsc{zh} \\
\midrule
\textbf{\small	Baselines} \\
\textsc{\footnotesize	 TOWER-V2 7B } & 69.4 & - & 79.5 & - & - & - & - & 78.5 & - & 82.1 & 75.6 & - & - \\
\textsc{\footnotesize MadLad400 7B } & 78.8 & 79.3 & 76.7 & - & 78.3 & 70.4 & 70.4 & 70.4 & 79.5 & 77.1 & 79.4 & 79.3 & 69.5\\
\textsc{\footnotesize NLLB 3.3B} & 75.5 & 76.0 & 75.3 & - & 74.5 & 69.0 & 70.9 & 66.9 & 76.6 & 79.0 & 73.5 & 75.1 & 60.5 \\
\cdashlinelr{1-16}
\textsc{\footnotesize Salamandra\textbf{TA}{\footnotesize 2B}} \\
\textsc{\tiny BASE {\tiny + CPT-v1} } & \textcolor{gray}{77.2} & \textcolor{gray}{77.3} & \textcolor{gray}{76.6} & \textcolor{gray}{-} & \textcolor{gray}{67.0} & \textcolor{gray}{-} & \textcolor{gray}{-} & \textcolor{gray}{-} & \textcolor{gray}{-} & \textcolor{gray}{-} & \textcolor{gray}{77.4} & \textcolor{gray}{76.0} & \textcolor{gray}{-} \\
\quad \textsc{\scriptsize + Instruct-v1} & 78.5 & 77.7 & 77.8 & - & 75.9 & - & - & - & - & - & 75.0 & 77.1 & - \\
\quad \quad \textsc{\scriptsize + TRR} & 83.4 & 85.1 & 82.4 & - & 81.6 & - & - & - & - & - & 81.4 & 82.6 & - \\
\quad \quad \textsc{\scriptsize + MBR} & 81.2 & 82.6 & 80.3 & - & 79.4 & - & - & - & - & - & 78.5 & 80.2 & - \\

\cdashlinelr{1-16}
\textsc{\footnotesize Salamandra\textbf{TA}{\footnotesize 7B}} \\
\textsc{\tiny BASE {\tiny + CPT-v1} } & \textcolor{gray}{77.8} & \textcolor{gray}{77.1} & \textcolor{gray}{77.2} & \textcolor{gray}{-} & \textcolor{gray}{75.6} & \textcolor{gray}{-} & \textcolor{gray}{-} & \textcolor{gray}{-} & \textcolor{gray}{-} & \textcolor{gray}{-} & \textcolor{gray}{78.0} & \textcolor{gray}{79.2} & \textcolor{gray}{-}
 \\
\quad \textsc{\scriptsize + Instruct-v1} & 81.3 & 82.6 & 80.4 & - & 79.9 & - & - & - & - & - & 78.5 & 80.0 & - \\
\quad \quad \textsc{\scriptsize + TRR} & 84.2 & 86.0 & 83.1 & - & 82.6 & - & - & - & - & - & 81.7 & 83.2 & - \\
\quad \quad \textsc{\scriptsize + MBR} & 82.5 & 83.8 & 81.3 & - & 80.8 & - & - & - & - & - & 79.3 & 81.0 & - \\
\cdashlinelr{1-16}
\textsc{\footnotesize Salamandra\textbf{TA}}{\scriptsize \textsc{-v2}} \\
\textsc{\scriptsize Base {\tiny + CPT-v1} {\tiny + CPT-v2} } & \textcolor{gray}{77.4} & \textcolor{gray}{76.9} & \textcolor{gray}{76.9} & \textcolor{gray}{78.2} & \textcolor{gray}{74.6} & \textcolor{gray}{69.2} & \textcolor{gray}{72.8} & \textcolor{gray}{73.7} & \textcolor{gray}{76.8} & \textcolor{gray}{75.9} & \textcolor{gray}{78.5} & \textcolor{gray}{78.7} & \textcolor{gray}{71.8}  \\
\quad \textsc{\scriptsize + Instruct-v2} & 80.2 & 81.6 & 79.7 & 82.7 & 79.9 & 75.2 & 68.8 & 78.7 & 80.6 & 79.3 & 77.7 & 78.4 & 70.1  \\
\quad \quad \textsc{\scriptsize + TRR} & 84.0 & 86.0 & 83.0 & 85.5 & 82.7 & 79.8 & 74.1 & 81.5 & 84.0 & 83.1 & 81.6 & 82.8 & 75.7  \\
\quad \quad \textsc{\scriptsize + MBR} & 82.0 & 83.9 & 81.1 & 83.9 & 80.6 & 76.9 & 71.1 & 80.0 & 82.3 & 81.1 & 78.9 & 80.3 & 71.7 \\

\bottomrule
\end{tabular}
\caption{\textsc{Comet-Kiwi} scores on the \wmtplusplus\ test set, comparing our \salamandraTA\ models against several strong baselines. We show the performance at each stage of our method: from the continually pre-trained base models (scores in gray), to the instruction-tuned models.}
\end{table*}

\begin{table*}
\centering
\setlength{\tabcolsep}{4.75pt}
\footnotesize
\renewcommand{\arraystretch}{1.3}
\begin{tabular}{lcccccccccccccccccc}
\toprule
&  \multicolumn{7}{c}{\textsc{comet}} &  \multicolumn{7}{c}{\textsc{metricX}} \\
& \textsc{de} & \textsc{el} & \textsc{it} & \textsc{lt} & \textsc{ro} & \textsc{sr} & \textsc{sv} & \textsc{de} & \textsc{el} & \textsc{it} & \textsc{lt} & \textsc{ro} & \textsc{sr} & \textsc{sv} \\
\midrule
\textsc{\footnotesize Salamandra\textbf{TA}{\footnotesize 2B}} \\
\textsc{\tiny BASE {\tiny + CPT-v1 } } & \\
\textsc{{\tiny + Instruct-v1} } & 76.6 & 83.5 & 78.6 & 79.7 & 80.3 & 75.3 & 80.9 & 2.31 & 4.10 & 4.03 & 5.20 & 4.22 & 6.18 & 3.28  \\
\quad \textsc{\scriptsize + TRR} & 80.6 & 85.7 & 82.2 & 83.7 & 84.1 & 80.8 & 84.5 & 1.63 & 3.37 & 2.62 & 3.85 & 3.02 & 4.53 & 2.25  \\
\quad \textsc{\scriptsize + MBR} & 81.9 & 86.6 & 83.4 & 85.1 & 85.0 & 81.5 & 85.3 & 1.60 & 3.39 & 2.69 & 3.84 & 3.08 & 4.71 & 2.33  \\
\cdashlinelr{1-16}
\textsc{\footnotesize Salamandra\textbf{TA}{\footnotesize 7B}} \\
\textsc{\tiny BASE {\tiny + CPT-v1 } } & \\
\textsc{{\tiny + Instruct-v1} } & 80.6 & 86.0 & 82.2 & 83.1 & 82.8 & 79.8 & 84.4 & 1.75 & 3.35 & 2.78 & 3.81 & 3.47 & 4.32 & 2.47  \\
\quad \textsc{\scriptsize + TRR} & 82.0 & 86.5 & 83.2 & 85.5 & 85.4 & 82.4 & 85.7  & 1.40 & 2.91 & 2.26 & 3.02 & 2.46 & 3.53 & 1.81 \\
\quad \textsc{\scriptsize + MBR} & 83.3 & 87.6 & 84.5 & 86.6 & 86.6 & 83.6 & 86.6 & 1.37 & 2.85 & 2.30 & 2.84 & 2.50 & 3.60 & 1.91  \\
\bottomrule
\end{tabular}
\caption{\textsc{comet} and \textsc{MetricX} scores for the WMT-Multilingual Sub-Task (English to seven target languages) on the \wmtplusplus\ test set. Results are shown for the instruction-tuned \salamandraTA{} 2B and 7B models, with and without post-decoding strategies (MBR and TRR).}\label{app:table_multilingual}
\end{table*}

\end{document}